\documentclass[preprint,12pt,review,authoryear]{elsarticle}

\usepackage{amssymb}
\usepackage{amsmath}
\usepackage{amsthm}
\usepackage{natbib}
\usepackage[table]{xcolor}
\usepackage{CJK}
\usepackage{tabularx}
\usepackage{subcaption}
\usepackage{siunitx}
\usepackage{tcolorbox}
\usepackage{fvextra}
\usepackage{adjustbox}
\usepackage{microtype}
\usepackage{inconsolata}
\usepackage{arydshln}
\usepackage{dirtytalk}
\usepackage{rotating}
\usepackage{fancyvrb}
\usepackage{graphicx}
\usepackage{makecell}
\usepackage{booktabs}
\usepackage{array}
\usepackage{lscape}
\usepackage{longtable}
\usepackage{multirow}
\usepackage{bm}
\usepackage[utf8]{inputenc}
\usepackage{algorithm}
\usepackage{algpseudocode}
\usepackage{enumitem}
\usepackage{graphicx}
\usepackage{pifont}
\usepackage{times}
\usepackage{listings}
\usepackage[utf8]{inputenc}
\usepackage{listingsutf8}
\usepackage{hyperref}
\usepackage{longtable}
\usepackage{makecell}
\usepackage{booktabs}
\usepackage[margin=1in]{geometry}
\usepackage{setspace}
\usepackage{pdflscape}
\usepackage{tcolorbox}
\usepackage{tabularx}
\usepackage{subcaption} 

\definecolor{mygreen}{RGB}{55,126,34}

\lstdefinestyle{json}{
    basicstyle=\ttfamily\footnotesize,  
    showstringspaces=false,
    breaklines=true,
    frame=single,
    backgroundcolor=\color{white},
    numbers=left,
    numberstyle=\tiny\color{gray},
    stepnumber=1,
    numbersep=5pt,
    keywordstyle=\color{mygreen},
    morekeywords={origin_id, origin_label, origin_text, generated_text, generated_tone, generated_label, topic_id, topic_words, doc_1_id, doc_1_label, doc_1_text, doc_2_id, doc_2_label, doc_2_text, generated_text_t01, generated_text_t015, origin_title, generated_text_glm4}
}

\journal{Decision Support Systems}

\begin{document}

\begin{frontmatter}



\title{MegaFake: A Theory-Driven Dataset of Fake News Generated by Large Language Models} 

\author[aff1]{Lionel Z. WANG}
\ead{lionel-z.wang@connect.polyu.hk}

\author[aff1]{Ka Chung NG\corref{cor1}}
\ead{kc-boris.ng@polyu.edu.hk}



\author[aff2]{Yiming MA}
\ead{yiming_ma2023@163.com}

\author[aff3,aff1]{Wenqi FAN}
\ead{wenqi.fan@polyu.edu.hk}

\cortext[cor1]{Corresponding author: Ka Chung NG.}

\affiliation[aff1]{organization={Department of Management and Marketing},
  addressline={Faculty of Business}, 
  city={The Hong Kong Polytechnic University}}
  
  
\affiliation[aff2]{addressline={Faculty of Computing}, 
  city={Harbin Institute of Technology}}
  
\affiliation[aff3]{organization={Department of Computing},
  addressline={Faculty of Computer and Mathematical Sciences}, 
  city={The Hong Kong Polytechnic University}}





\begin{abstract}
Fake news significantly influences decision-making processes by misleading individuals, organizations, and even governments. Large language models (LLMs), as part of generative AI, can amplify this problem by generating highly convincing fake news at scale, posing a significant threat to online information integrity. Therefore, understanding the motivations and mechanisms behind fake news generated by LLMs is crucial for effective detection and governance. In this study, we develop the LLM-Fake Theory, a theoretical framework that integrates various social psychology theories to explain machine-generated deception. Guided by this framework, we design an innovative prompt engineering pipeline that automates fake news generation using LLMs, eliminating manual annotation needs. Utilizing this pipeline, we create a theoretically informed \underline{M}achin\underline{e}-\underline{g}ener\underline{a}ted \underline{Fake} news dataset, MegaFake, derived from FakeNewsNet. Through extensive experiments with MegaFake, we advance both theoretical understanding of human-machine deception mechanisms and practical approaches to fake news detection in the LLM era.
\end{abstract}




\end{frontmatter}

\section{Introduction}\label{intro}
In an era where information drives critical decisions, the proliferation of fake news poses a significant challenge to individuals, organizations, and governments \citep{ng2021effect, ng2023augmenting}. Decision-making processes heavily rely on the accuracy and reliability of the information consumed. However, the recent development of large language models (LLMs), amplified the complexity of this issue by enabling malicious actors to produce extensive volumes of machine-generated fake news \citep{vykopal-etal-2024-disinformation, chen2024llmgenerated}. This raises a pivotal decision-making problem for platform owners and policymakers about \textbf{how to effectively identify machine-generated fake news in order to improve online content quality and support content governance}.

Addressing this issue requires tools and resources that enable a deeper understanding of the characteristics, patterns, and risks of fake news in the age of LLMs. Although a growing body of research has emerged to tackle machine-generated fake news or deepfake text, existing studies exhibit critical limitations that constrain their practical applicability.
First, many recent works focus on \textbf{isolated technical aspects or specific attack vectors}, overlooking the holistic nature of deception. For instance, \citet{wang2025prompt} examine prompt-induced linguistic fingerprints, while \citet{wu2024sheepdog} address style-based attacks. Similarly, \citet{das2025fake} and \citet{su2024adapting} investigate detection performance following ``LLM laundering'' or paraphrasing. While valuable, these studies primarily treat fake news as a text classification problem centered on specific linguistic features, and they lack a unifying theoretical framework to explain the underlying socio-psychological deception mechanisms. Second, existing datasets are often constrained by \textbf{narrow domains or limited scale}. For example, \citet{hong2024leveraging} and \citet{sun2024exploring} focus on domain-specific topics such as COVID-19 or healthcare misinformation. Although \citet{ayoobi2024seeing} attempt to enhance human skepticism, the datasets used in these studies are typically small or domain-specific, which limits the generalizability of detection models across diverse topics and deceptive scenarios. 
Third, and perhaps most critically, the \textbf{generation strategies} adopted in current literature rely heavily on intuitive or simple prompting techniques without deep theoretical grounding. Studies by \citet{huang2024fakegpt}, \citet{chen2024llmgenerated}, and \citet{vykopal-etal-2024-disinformation} have explored the capabilities of LLMs to generate fake news, yet they often lack a systematic categorization of deceptive intents. As a result, the resulting datasets fail to capture the nuanced spectrum of deception---from subtle exaggeration to complete fabrication---that characterizes real-world disinformation campaigns. Consequently, there is a scarcity of resources that organize machine-generated deception according to established social-psychological theories, which in turn hinders the development of robust detection systems capable of understanding \textit{why} and \textit{how} content is designed to deceive.

To bridge these gaps, it is imperative to develop a large-scale, diverse dataset that is explicitly guided by established theories in social psychology. Such a resource is essential not only for elucidating the deceptive motivations and mechanisms underlying machine-generated fake news, but also for advancing research by providing a robust benchmark for future detection models.

In this paper, we introduce the \underline{M}achin\underline{e}-\underline{g}ener\underline{a}ted \underline{Fake} news dataset, named \textbf{MegaFake}, a dataset comprising four types of fake news and two types of legitimate news generated by LLMs. Built on the GossipCop and PolitiFact datasets from FakeNewsNet~\citep{shu2020fakenewsnet}, MegaFake is generated using both the General Language Model (GLM)~\citep{glm2024chatglm} and Llama~\citep{touvron2023llama}. For each model, the dataset includes 129,085 fake news instances and 41,706 legitimate news instances. This dataset is one of the first large-scale, publicly available collections of machine-generated fake news. It is grounded in our theoretical framework, \textbf{LLM-Fake Theory}, which guides the generation process. We have conducted several analyses to demonstrate the high quality of the dataset. Importantly, we have provided an ethical statement in the \href{https://osf.io/x7wrs/overview?view_only=cff5842af7b54b168d0479df985a6127}{Online Data Repository} that details the procedures and ethical considerations involved in the creation and use of the MegaFake dataset.

We conduct a series of experiments to demonstrate the value of MegaFake in supporting fake news detection, analysis, and governance: 1) \textbf{Deception Detection Analysis}: We compare the performance of various LLMs and deep learning models in detecting both human- and machine-generated fake news using the newly established MegaFake dataset. These insights inform the development of more robust fake news detection models, helping to safeguard the quality of online content. 2) \textbf{Machine-Generated Content Differentiation}: We introduce new benchmarks for detecting diverse types of machine-generated content. These benchmarks facilitate the advancement of detection methods, enabling more effective governance of online information, especially content produced by LLMs. 3) \textbf{Human-Machine Cross-Domain Experiments}: We conduct cross-domain experiments by training on human-generated fake news and testing on machine-generated content, and vice versa. The significant performance drops observed suggest distinct patterns in human and machine deception, highlighting the value of MegaFake for stimulating and advancing future research into the underlying mechanisms of both human- and machine-generated fake news. 4) \textbf{Generalization Capability Analysis}: We benchmark MegaFake against seven representative machine-generated fake news datasets under strictly controlled training sizes to assess intrinsic data quality. Models trained on MegaFake consistently achieve superior performance on an external test set, indicating higher information density and greater linguistic diversity. These results validate MegaFake's effectiveness in enhancing model generalization. 5) \textbf{Data Scale Sensitivity}: We investigate the scaling law of fake news detection by evaluating model performance across training set sizes ranging from 1k to 60k samples. Our findings reveal a substantial performance gain up to around 20k samples, showing that existing small-scale datasets are insufficient for optimal training and highlighting the necessity of MegaFake’s large scale to enhance detection performance.

Our research makes several theoretical and practical contributions to the literature. First, we add to the fake news literature by developing the LLM-Fake Theory, which identifies four methods for generating fake news using LLMs, each grounded in established social psychology theories \citep{chatman1975towards, chen2020linguistic, hinojosa2017review, petty2011elaboration}. This framework explicates the underlying deceptive mechanisms and motivations, focusing specifically on intentional fake news \citep{chen2024llmgenerated}, rather than unintentional misinformation or \say{hallucinations} \citep{10.1145/3703155, zhang2023siren}. We also explore two strategies for generating legitimate news using LLMs. Second, our MegaFake dataset and theory-guided generation pipeline provide essential resources for developing models targeting machine-generated content created by LLMs. Third, our experiments yield important insights into deepfake text detection and governance. Notably, we identify a critical gap in current detection systems, as significant performance degradation occurs when models trained on human-generated fake news are tested on machine-generated content and vice versa. This suggests the need for more robust and adaptable models capable of addressing both human and machine deception patterns, ultimately supporting more effective online content governance and informed decision-making.

\section{Related Work}\label{related work}
\subsection{Human-Generated Fake News and Detection}
The term ``fake news'' is widely used but fails to capture the full range of related issues. Major studies emphasize the \textbf{intentionality} behind fake news, focusing on the deliberate dissemination of false information and the governance of such content \citep{pennycook2021psychology, lazer2018science, zhang2020overview, gelfert2018fake}. \cite{murayama2021dataset} notes that definitions of fake news range from broad to narrow, while \cite{lazer2018science} define it as ``fabricated information that mimics news media content in form but not in organizational process or intent.''

Research on fake news detection heavily relies on the dataset construction for developing and evaluating detection models. Several datasets have been introduced; for example, the GossipCop and PolitiFact datasets \citep{shu2020fakenewsnet} provide substantial collections of labeled fake news articles along with associated social media posts. This resource has become a popular benchmark for training fake news detection models \citep{doi:10.1126/science.aap9559, wang-2017-liar, morita2025genaireading,watanabe2024comparing}.

Traditional approaches to fake news detection employ a combination of content-based and context-based features. Content-based methods analyze linguistic elements, such as syntax and semantics, to identify inconsistencies that may indicate falsehoods \citep{pan2018content, della2018automatic}. In contrast, context-based methods examine the propagation patterns of news articles, using social network analysis to detect anomalies \citep{10.1145/3137597.3137600, doi:10.1126/science.aap9559, shu2019beyond}. Recent advances in natural language processing have introduced sophisticated deep learning models for fake news detection \citep{kaliyar2021fakebert, ng2023augmenting}. Notably, several theory-driven methods \citep{lee2024explainable, zhang2022theory, zhou2020fake} have been proposed to enhance fake news detection through explainable approaches, providing greater transparency and interpretability in the detection process.

\subsection{Methods for Generating Fake News with LLMs}\label{sec:generation_methods}
The emergence of LLMs has introduced new paradigms for creating deceptive content. To better understand the landscape of machine-generated fake news or deepfake text, we categorize existing generation methods into three primary streams: \textit{Direct Fabrication}, \textit{Style-Based Manipulation}, and \textit{Context-Conditional Generation}.

\textbf{Direct Fabrication.} 
The most straightforward approach involves prompting LLMs to generate fake news from scratch, typically by providing a target topic or a specific fake news narrative. Studies such as \citet{vykopal-etal-2024-disinformation} and \citet{chen2024llmgenerated} have demonstrated that LLMs can generate convincing news articles that align with harmful disinformation narratives when explicitly instructed. Similarly, \citet{huang2024fakegpt} employed various prompting methods to fabricate news samples, showing that LLMs can produce high-quality fake content that mimics human reasoning. However, these methods often rely on intuitive, open-ended prompts, which can yield generic content that lacks the nuanced deceptive intent characteristic of real-world disinformation campaigns.

\textbf{Style-Based Manipulation.} 
A second category focuses on altering the stylistic presentation of news rather than fabricating events outright. This often involves rewriting or ``laundering'' existing news articles. \citet{wu2024sheepdog} showed that LLMs can mimic the writing style of trustworthy news sources to camouflage fake content, significantly degrading the performance of style-based detectors. \citet{das2025fake} further explored ``LLM laundering,'' in which LLMs paraphrase fake news to evade detection while preserving the original semantic meaning. Additionally, \citet{lucas-etal-2023-fighting} used perturbation-based prompts to synthesize deceptive content. Although effective at mimicking linguistic patterns, these methods primarily target the \textit{presentation} of fake news rather than the \textit{creation} of new false facts or narratives.

\textbf{Context-Conditional Generation.} 
More recent approaches attempt to control the generation process more granularly through advanced prompting techniques. For instance, \citet{sun2024exploring} propose a variational-autoencoder-like prompt to generate fake news in the healthcare domain, aiming to maintain contextual consistency without human intervention. \citet{wang2025prompt} analyzed prompt-induced linguistic fingerprints, showing that specific prompting strategies lead to distinct probability shifts in the generated text.

While these studies showcase the capabilities of LLMs, current generation methods remain fragmented. Existing work typically adopts only one of the above strategies in isolation, failing to capture the diverse spectrum of deception. Moreover, there is no unified theoretical framework that categorizes these methods according to the underlying psychological and cognitive mechanisms of deception. To address this gap, we develop the LLM-Fake Theory and introduce the MegaFake dataset, providing a theory-driven and enriched data foundation to significantly advance research in this domain.

\subsection{Machine-Generated Fake News Detection}\label{sec:detection_datasets}
Researchers have proposed various methods to distinguish machine-generated fake news from human-written content. \citet{jiang2024disinformation} and \citet{su2024adapting} conducted comprehensive evaluations of existing detectors, revealing that models trained on human-written news often fail to generalize to machine-generated content. To address this, \citet{wu2024sheepdog} introduced style-robust detectors, while \citet{wang2024explainable} and \citet{wang2025prompt} focused on extracting linguistic fingerprints and explanation-based features. Furthermore, studies such as \citet{hong2024leveraging} and \citet{ayoobi2024seeing} have explored the dual role of LLMs, employing them not only as generators but also as evaluators to support both human skepticism and automated detection.

Despite these advancements, the development of robust detection models is hindered by limitations in existing datasets:
1) \textbf{Limited Methodological Diversity}: Most datasets are constructed using a single generation method (e.g., only paraphrasing or only direct generation), which fails to capture the multifaceted nature of machine-generated fake news. 
2) \textbf{Domain Restrictions}: Many datasets focus on narrow topics, such as COVID-19 misinformation \citep{hong2024leveraging, sun2024exploring}, which constrains the generalizability of models trained on them. 
3) \textbf{Insufficient Scale}: Datasets such as those used in \citet{ayoobi2024seeing} often lack the volume needed to train large, data-hungry deep learning models.

To bridge these gaps, we introduce the MegaFake dataset, a systematically diverse and theoretically grounded resource that offers a broader spectrum of deception strategies than prior works. MegaFake is constructed using the four distinct generation methods defined in our LLM-Fake Theory. It provides a richer label structure, including four specific types of fake news alongside legitimate machine-generated news. With over 120,000 articles drawn from a wide range of sources and topics, the MegaFake dataset substantially surpasses the size of existing domain-specific datasets.

\section{LLM-Fake Theory}\label{Theory}
Drawing on an extensive review of the literature \citep{su2024adapting,chen2024llmgenerated,huang2024fakegpt,jiang2024disinformation,lucas-etal-2023-fighting,vykopal-etal-2024-disinformation, zhou2023synthetic} on generating and detecting machine-generated fake news, we integrate insights from various social psychology theories \citep{chatman1975towards, chen2020linguistic, hinojosa2017review, petty2011elaboration} to develop the LLM-Fake Theory. This theoretical framework is designed to synthesize the social psychological rationale, deceptive motivations, and mechanisms behind the creation of fake news by LLMs. This theory categorizes machine-generated content into four types of fake content and two types of legitimate content. These categories are meticulously defined to encompass a wide range of news scenarios, ranging from the ethical enhancement of writing and language through LLMs \citep{luo2023chatgpt,ROSSI2024102749,susarla2023janus} to the malicious generation of fake news \citep{schuster2020limitations,wu2024sheepdog,zhou2023synthetic}. Notably, it is crucial to distinguish the intentional, prompt-guided machine deception that is the focus of this study from unintentional LLM hallucination. While both can result in falsehoods, the former follows structured, psychologically motivated patterns to mislead, whereas the latter refers to the model's spontaneous generation of factually incorrect or nonsensical content. \textbf{Tables \ref{tab:machine_fake1}} and \textbf{\ref{tab:machine_legitimate1}} present our theory-guided prompts for generating different types of fake and legitimate news using LLMs \cite{liu-etal-2025-sara}. Below, we describe the theoretical foundations underlying each generation approach.

\begin{table*}[ht]
\scriptsize
\setlength{\tabcolsep}{11pt}
\caption{Four types of machine-generated fake news.}
\resizebox{\textwidth}{!}{ 
    \begin{tabular}{p{2.5cm}p{3cm}p{8cm}} 
        \toprule [1.2pt]
        \specialrule{0em}{1pt}{1pt}
        
        News Type & Definition & Prompt Template Example \\ 
        
        \specialrule{0em}{1pt}{1pt}
        \hline 
        \specialrule{0em}{0pt}{0pt}
        
        \makecell[{{p{2.5cm}}}]{
        \renewcommand\arraystretch{0.1}
        Sheep’s Clothing (Explained by Linguistic Signaling Theory \citep{chen2020linguistic})}  & 
        \makecell[{{p{3cm}}}]{Utilize an LLM to rephrase legitimate news in the style of tabloids or fake news in the style of mainstream sources.} & 
        \makecell[{{p{8cm}}}]{
        For a human-generated fake news article as an input: \\[2pt]
        1. Rewrite the following news article in an objective and professional tone without changing the content and meaning while keeping a similar length. \\[2pt]
        {[}fake news article{]} \\[2pt]
        2. Rewrite the following news article in a neutral tone without changing the content and meaning while keeping a similar length. \\[2pt]
        {[}fake news article{]} \\[2pt]
        For a human-generated legitimate news article as an input: \\[2pt]
        3. Rewrite the following news article in an emotionally triggering tone without changing the content and meaning while keeping a similar length. \\[2pt]
        {[}legitimate news article{]} \\[2pt]
        4. Rewrite the following news article in a sensational tone without changing the content and meaning while keeping a similar length. \\[2pt]
        {[}legitimate news article{]}} \\

        \specialrule{0em}{2pt}{2pt}
        \midrule
        \specialrule{0em}{2pt}{2pt}
        
        \makecell[{{p{2.5cm}}}]{Content Manipulation (Explained by Elaboration Likelihood Model \citep{petty2011elaboration}))} & 
        \makecell[{{p{3cm}}}]{Utilize an LLM to manipulate human-generated legitimate news content by modifying multiple attributes.} & 
        \makecell[{{p{8cm}}}]{For a human-generated legitimate news article as an input: \\[2pt]
        Modify the attributes, such as events, statements, actions, and numerical quantities, in the following news article by minimizing the editing and keeping the same language style and a similar length. \\[2pt]
        {[}legitimate news article{]}} \\
        
        \specialrule{0em}{2pt}{2pt}
        \midrule
        \specialrule{0em}{2pt}{2pt}
        
        \makecell[{{p{2.5cm}}}]{Information Blending (Explained by Cognitive Dissonance Theory \citep{hinojosa2017review})} & 
        \makecell[{{p{3cm}}}]{Utilize an LLM to integrate fake and legitimate news for creating new fake content.} & 
        \makecell[{{p{8cm}}}]{For a pair of human-generated fake and legitimate news articles on comparable subjects as an input: \\[2pt]
        Amalgamate the following two news articles into a new and cohesive article while keeping a similar length. \\[2pt]
        {[}fake news article{]}, {[}legitimate news article{]}}\\
        
        \specialrule{0em}{2pt}{2pt}
        \midrule
        \specialrule{0em}{2pt}{2pt}
        
        \makecell[{{p{2.5cm}}}]{Narrative Generation (Explained by Narrative Theory \citep{chatman1975towards})} & 
        \makecell[{{p{3cm}}}]{Utilize an LLM to generate fake news from a certain message.} & 
        \makecell[{{p{8cm}}}]{For either a human-generated fake or legitimate news article as an input: \\[2pt]
        Write a news article based on the following message and return the body content only. \\[2pt]
        {[}news article title{]}} \\
        
    \bottomrule [1.2pt]
    \end{tabular}}
    \label{tab:machine_fake1} 
\end{table*}
\begin{table*}[ht]
\scriptsize
\setlength{\tabcolsep}{11pt}
\renewcommand{\arraystretch}{0.1}
\caption{Two types of machine-generated legitimate news.}
\resizebox{\textwidth}{!}{ 
    \begin{tabular}{p{2.5cm}p{3cm}p{8cm}} 
        \toprule [1.2pt]
        \specialrule{0em}{2pt}{2pt}
        
        News Type & Definition & Prompt Template Example \\ 
        
        \specialrule{0em}{1pt}{1pt}
        \hline 
        \specialrule{0em}{1pt}{1pt}
        
        \makecell[{{p{2.5cm}}}]{Writing Enhancement \citep{ROSSI2024102749,susarla2023janus,wu2024sheepdog,yang2023chatgpt}} & 
        \makecell[{{p{3cm}}}]{Utilize an LLM to polish legitimate news while preserving the original information and context.}  & 
        \makecell[{{p{8cm}}}]{For a human-generated legitimate news article as an input: \\ \\
        Polish the following news article to make it more objective and professional. Do not change the original meaning. Do not add extra information or delete certain information. \\ \\
        {[}legitimate news article{]} }\\
        
        \specialrule{0em}{2pt}{2pt}
        \midrule
        \specialrule{0em}{2pt}{2pt}
               
        \makecell[{{p{2.5cm}}}]{News Summarization \citep{luo2023chatgpt,wu2024sheepdog, zhang2023siren}} & 
        \makecell[{{p{3cm}}}]{Utilize an LLM to condense various legitimate news into a synthesized, credible news summary.} &
        \makecell[{{p{8cm}}}]{For a pair of human-generated legitimate news articles on comparable subjects as an input: \\ \\
        Summarize the following two articles into a single, cohesive article, ensuring minimal loss of information without significantly shortening the overall length. \\ \\
        {[}legitimate news article 1{]}, {[}legitimate news article 2{]}}\\
        
    \bottomrule [1.2pt]
    \end{tabular}}
    \label{tab:machine_legitimate1} 
\end{table*}

\subsection{Sheep’s Clothing: Style-Based Machine-Generated Fake News}
This strategy use an LLM to alter the style of news articles \citep{wu2024sheepdog}, shaping audience perceptions and beliefs through mechanisms explained by linguistic signaling theory \citep{chen2020linguistic}. According to this theory, language style serves as a signal that conveys qualities about the communicator, influencing how the message is received. By mimicking the style of reputable media, LLM-generated fake news can appear more credible and professional, misleading readers about its authenticity. Conversely, adopting styles associated with less credible sources, such as sensationalism or emotional language, can reduce the perceived trustworthiness of the content. Linguistic signaling theory thus provides a framework for understanding how stylistic changes in news articles manipulate audience perceptions, enabling styled fake news to deceive effectively.

\subsection{Content Manipulation: Content-Based Machine-Generated Fake News}
The second strategy involves using an LLM to alter legitimate news by modifying attributes like events and numbers. The elaboration likelihood model \citep{petty2011elaboration} explains how people process such manipulative communications. It outlines two main ways of processing information \citep{bhattacherjee2006influence}. On one hand, the central route involves deliberate and thoughtful consideration of the information's true merits when people are motivated and able to think critically. This deep analysis allows people to notice and assess changes in the content. On the other hand, the peripheral route involves superficial processing based on external cues like source credibility rather than the content’s actual quality. It is common when people are either unmotivated or unable to critically analyze the information. When LLMs subtly alter facts in news content, these changes might not be obvious, especially to those relying on the peripheral route, such as source credibility. Therefore, if the news looks professional and comes from a seemingly trustworthy source, readers may accept the altered information as true without thorough scrutiny. In a digital environment where users often rapidly skim content, there is a tendency to rely more on this peripheral processing, making subtly manipulated news more effective.

\subsection{Information Blending: Integration-Based Machine-Generated Fake News}
The third type of fake news generated by an LLM involves blending legitimate and fake news to create new, misleading content. This manipulation technique can be explained by the cognitive dissonance theory \citep{hinojosa2017review}, which posits that people experience discomfort when faced with contradictory beliefs or information. In mixed-content fake news, readers may recognize some elements as true (aligning with their existing beliefs or knowledge) while perceiving others as potentially false. To alleviate this discomfort, known as cognitive dissonance, they may adjust their perceptions or dismiss inconsistencies, leading to the erroneous acceptance of the entire article as true \citep{moravec2018fake}. This effect is particularly potent if the true elements are more prominent or if the false elements align with readers' preconceived notions or biases \citep{moravec2018fake}. By skillfully mixing truth with falsehoods, LLM-generated fake news exploits a natural cognitive bias: the human tendency to avoid the discomfort of conflicting information. As a result, such content becomes not only more pleasant but also harder to critically evaluate and reject. To generate mixed-content news using an LLM, we employ a topic model \citep{alambo2020topic,vayansky2020review} to identify pairs of legitimate and fake news articles with similar topics and content for effective blending.

\subsection{Narrative Generation: Story-Based Machine-Generated Fake News} 
The final fake news type involves creating entirely fictional stories using an LLM. This process can be analyzed through narrative theory \citep{chatman1975towards}, which offers a framework for understanding how fabricated news manipulates audience perceptions and beliefs. Narrative theory divides a story into two main components: story and discourse. The story consists of the core elements of the narrative, including the sequence of events (actions and happenings) and the existents (characters and settings) \citep{chatman1975towards,schuster2020limitations}. LLMs create and arrange these elements, such as inventing a political event or fictional characters. The discourse, on the other hand, refers to how the story is told---such as the choice of language, narrative techniques, and the structuring of events \citep{chatman1975towards,schuster2020limitations}. LLMs choose how to phrase and structure the narrative, strategically emphasizing or omitting details to craft an impactful discourse. By generating detailed and plausible stories, LLMs make fake news difficult to distinguish from legitimate news, especially when the content aligns with popular beliefs or existing narratives.

\subsection{Writing Enhancement: Improvement-Based Machine-Generated Legitimate News}
This type of legitimate news aims to enhance the writing of human-generated legitimate news using an LLM. This involves polishing the content while preserving the original information and context. Existing studies suggest that LLMs can support the writing process, such as improving prose and correcting grammar \citep{ROSSI2024102749,susarla2023janus,wu2024sheepdog,yang2023chatgpt}. Thus, we expect that LLMs will similarly be useful in enhancing the writing and word usage of news articles. To this end, a prompt is specifically designed to facilitate the generation of objective and professional legitimate news articles, emphasizing factual accuracy, coherence, and stylistic quality.

\subsection{News Summarization: Summary-Based Machine-Generated Legitimate News}
The second type employs an LLM to synthesize news articles with similar content and topics. Research indicates that LLMs are transforming how we consume news by efficiently summarizing lengthy articles into concise overviews \citep{luo2023chatgpt,wu2024sheepdog,zhang2024benchmarking}. These summaries allow individuals to quickly grasp essential information, facilitating more efficient and informed decision-making. By reducing cognitive load and making news consumption quicker, LLMs help users stay informed without feeling overwhelmed, adapting to the increasing demands of our information-rich world \citep{xu2023chatgpt}. In this regard, we use an LLM to summarize pairs of human-generated legitimate news articles matched by a topic model \citep{alambo2020topic,vayansky2020review}, ensuring minimal information loss and maintaining sufficient length for effective model training.

\section{Dataset Construction}\label{Dataset} 

\subsection{Generation Pipeline}
\begin{figure*}[!ht]
    \centering
    \includegraphics[width=0.9\textwidth]{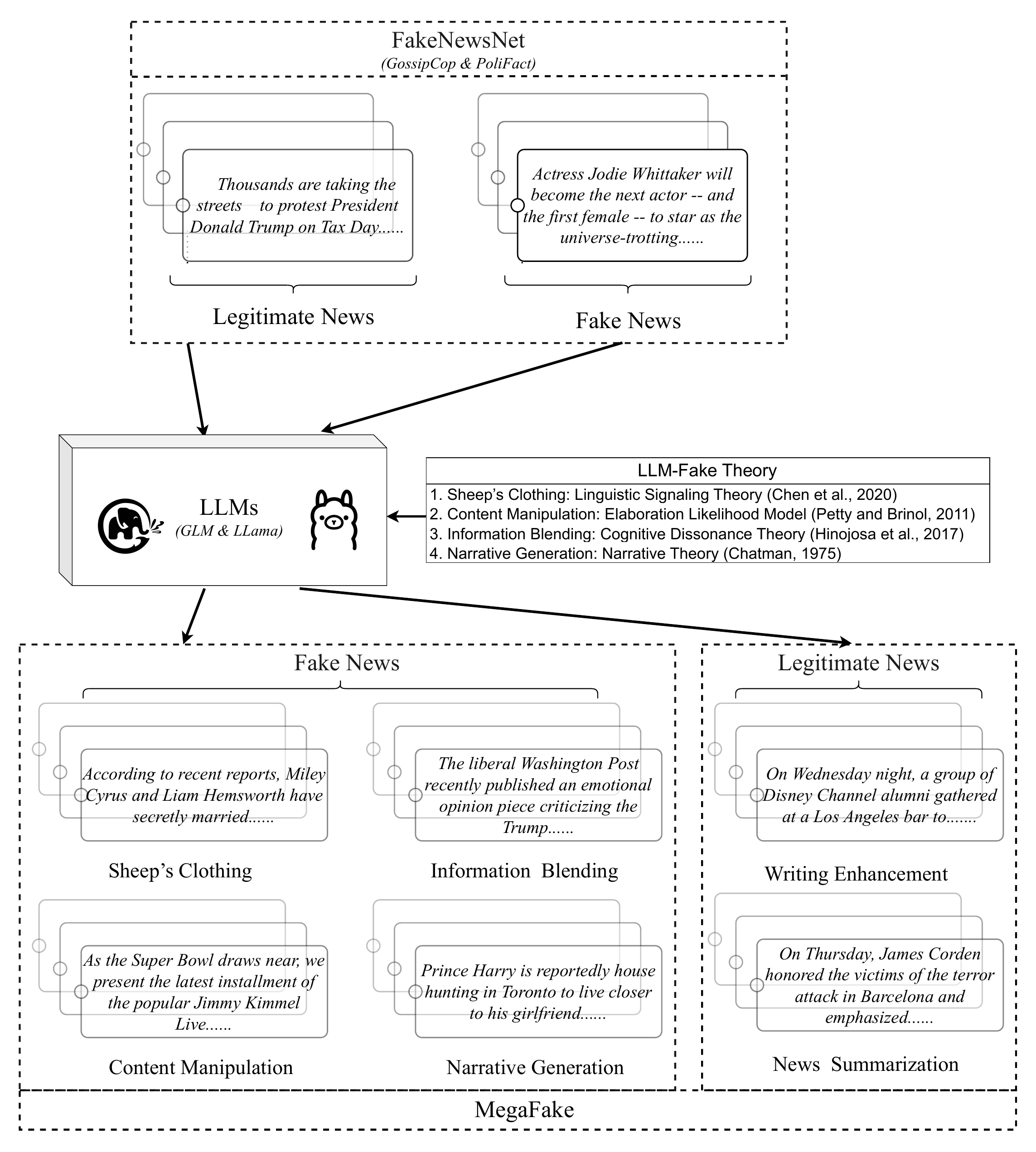}
    \caption{Generation pipeline for the MegaFake dataset.}
    \label{fig:pipeline}
\end{figure*}

To construct the MegaFake dataset, we first input the prompts, as detailed in \textbf{Tables \ref{tab:machine_fake1}} and \textbf{\ref{tab:machine_legitimate1}}, into the selected LLMs and compile their outputs. The resulting dataset encompasses diverse categories of generated content. \textbf{Figure \ref{fig:pipeline}} illustrates the generation pipeline for the MegaFake dataset. We apply the LLM-Fake Theory within this pipeline to guide the generation process. We begin by sourcing the GossipCop and PolitiFact datasets from the FakeNewsNet data repository \citep{shu2020fakenewsnet}. For LLM selection, we choose the General Language Model (GLM) \citep{glm2024chatglm} and Llama \citep{touvron2023llama}, both of which have showed superior performance compared to models like GPT \citep{achiam2023gpt} across equivalent model sizes and datasets \citep{du2023glm}. We provide further details in the \href{https://osf.io/x7wrs/overview?view_only=cff5842af7b54b168d0479df985a6127}{Online Data Repository}.

\subsection{Dataset Quality Evaluation}
\subsubsection{Hallucination Analysis}

To distinguish intentional deception from unintended model errors, we employ the PURIFY framework \citep{lucas-etal-2023-fighting} to assess hallucination risks. The aim of this analysis is to quantify the impact of such unintentional artifacts, allowing us to determine whether the deceptive characteristics observed in MegaFake arise primarily from intentional, theory-guided design rather than from random model failures. While prior studies report hallucination rates as high as 38\% in GPT-3.5 \citep{shi-etal-2023-hallucination}. We apply PURIFY to evaluate two types of fake news---\say{sheep's clothing} and \say{information blending}---as well as all legitimate news types in MegaFake. Our evaluation reveals a significantly lower overall misalignment rate of 9.23\%. Specifically, we observe only 1.05\% logical misalignment and 11.37\% factual inconsistencies, two error modes typically driven by LLM hallucinations. These results suggest that the deceptive content in MegaFake is predominantly driven by our theoretical design rather than by model artifacts. A detailed analysis is provided in the \href{https://osf.io/x7wrs/overview?view_only=cff5842af7b54b168d0479df985a6127}{Online Data Repository}.

\subsubsection{LLM Deception Comparison}
Next, we evaluate and compare the content generated by GLM and Llama by focusing on two key aspects: fluency and diversity. For fluency, we use perplexity (PPL), a standard language modeling metric, to assess the naturalness of the generated text. For diversity, we calculate Distinct-n, which measures the ratio of unique n-grams to the total number of n-grams in the dataset, and the Type-Token Ratio (TTR), which represents the proportion of unique words (types) to the total number of words (tokens). \textbf{Table \ref{tab:auto}} presents the comparison results based on these metrics. The Llama-generated MegaFake dataset exhibits lower perplexity, indicating higher fluency or naturalness. In contrast, the GLM-generated MegaFake dataset shows higher Distinct-1, Distinct-2, and TTR values, reflecting greater lexical and phrase diversity. These results suggest a trade-off: Llama-generated texts are more fluent, while GLM-generated texts are more diverse.

\begin{table}[!ht]
\centering
\caption{Comparison of content generated by GLM and Llama.}
\begin{tabular}{lcccc}
\toprule
Dataset & Perplexity $\downarrow$ & Distinct-1 $\uparrow$ & Distinct-2 $\uparrow$ & TTR $\uparrow$ \\
\midrule
GLM-Generated MegaFake & 13.52 & 0.66 & 0.54 & 0.14 \\
Llama-Generated MegaFake & 11.74 & 0.58 & 0.43 & 0.13 \\
\bottomrule
\multicolumn{5}{l}{\footnotesize$\downarrow$ indicates that lower values correspond to better performance, whereas $\uparrow$ indicates that higher values} \\
\multicolumn{5}{l}{\footnotesize correspond to better performance.}
\end{tabular}
\label{tab:auto}
\end{table}

\subsubsection{Human Evaluation}
We assess the quality of 1,708 samples (a random 1\% of the dataset) using two groups of annotators: 50 general annotators and ten trained university students. The evaluation demonstrates high consistency (average Cohen’s $\kappa=0.82$). Specifically, the general annotators rate 96.14\% of articles as fluent and 98.72\% as free from toxicity, while the student reviewers confirm that 89.16\% of samples meet high standards of quality and appropriateness. A detailed description of the annotation procedure is provided in the \href{https://osf.io/x7wrs/overview?view_only=cff5842af7b54b168d0479df985a6127}{Online Data Repository}.

\subsection{Dataset Statistics}

\textbf{Table~\ref{tab:statistics_news}} presents descriptive statistics for MegaFake. We report a range of linguistic features, including the average number of sentences per text (\textit{Sent Count}), the average number of words per text (\textit{Word Count}), the average sentence length in words, reflecting syntactic complexity (\textit{Sent Len (Words)}), the average sentence length in characters, indicating expression density (\textit{Sent Len (Chars)}), and the average word length, which serves as an indicator of lexical complexity (\textit{Word Len}). In addition, sentiment features are computed using VADER\footnote{\url{https://github.com/cjhutto/vaderSentiment}}, an open-source Python package specifically attuned to sentiments expressed in social media. These features include the overall sentiment score (\textit{Polarity}, ranging from -1 to +1, with negative values denoting negative sentiment) and its standard deviation (\textit{Polarity Std}, reflecting sentiment consistency).

\begin{table}[!ht]
\centering
\label{tab:megafake_comparison}
\caption{Descriptive statistics of the MegaFake dataset.}
\resizebox{\textwidth}{!}{%
\begin{tabular}{lrrrrrrrr}
\toprule
News Type & Sample & Sent Count & Word Count & Sent Len (Words) & Sent Len (Chars) & Word Len & Polarity & Polarity Std \\
\midrule
\multicolumn{9}{l}{\textbf{GLM-Generated Content Based on GossipCop}} \\
\midrule
Sheep's Clothing (Fake) & 15729 & 12.68 & 291.19 & 22.96 & 113.70 & 4.16 & 0.58 & 0.68 \\
Content Manipulation (Fake) & 11941 & 17.38 & 398.24 & 22.91 & 115.25 & 4.27 & 0.71 & 0.58 \\
Information Blending (Fake) & 2697 & 12.35 & 308.08 & 24.95 & 126.63 & 4.27 & 0.55 & 0.68 \\
Narrative Generation (Fake) & 15421 & 10.04 & 227.31 & 22.64 & 113.75 & 4.22 & 0.64 & 0.63 \\
Writing Enhancement (Legitimate) & 11945 & 9.83 & 229.95 & 23.38 & 118.26 & 4.27 & 0.52 & 0.67 \\
News Summarization (Legitimate) & 5926 & 10.40 & 263.48 & 25.33 & 129.09 & 4.29 & 0.54 & 0.70 \\
\midrule
\multicolumn{9}{l}{\textbf{GLM-Generated Content Based on PolitiFact}} \\
\midrule
Sheep's Clothing (Fake) & 7001 & 22.14 & 588.28 & 26.57 & 147.88 & 4.76 & 0.05 & 0.88 \\
Content Manipulation (Fake) & 2445 & 32.06 & 744.13 & 23.21 & 127.08 & 4.69 & 0.45 & 0.79 \\
Information Blending (Fake) & 2000 & 15.50 & 455.34 & 29.38 & 168.44 & 4.91 & -0.05 & 0.84 \\
Narrative Generation (Fake) & 7001 & 11.03 & 310.88 & 28.18 & 164.92 & 5.01 & 0.54 & 0.68 \\
Writing Enhancement (Legitimate) & 2445 & 22.43 & 614.17 & 27.38 & 154.83 & 4.85 & 0.32 & 0.82 \\
News Summarization (Legitimate) & 2000 & 18.38 & 569.80 & 30.99 & 177.29 & 4.90 & 0.26 & 0.84 \\
\midrule
\multicolumn{9}{l}{\textbf{Llama-Generated Content Based on GossipCop}} \\
\midrule
Sheep's Clothing (Fake) & 15729 & 14.53 & 359.08 & 24.71 & 123.57 & 4.21 & 0.49 & 0.78 \\
Content Manipulation (Fake) & 11945 & 18.81 & 472.27 & 25.10 & 121.09 & 4.04 & 0.65 & 0.65 \\
Information Blending (Fake) & 3000 & 15.66 & 415.29 & 26.52 & 136.33 & 4.32 & 0.80 & 0.50 \\
Narrative Generation (Fake) & 15729 & 13.54 & 348.91 & 25.76 & 132.60 & 4.34 & 0.75 & 0.58 \\
Writing Enhancement (Legitimate) & 11945 & 13.65 & 337.19 & 24.71 & 129.44 & 4.44 & 0.62 & 0.66 \\
News Summarization (Legitimate) & 3000 & 15.82 & 440.05 & 27.81 & 144.25 & 4.37 & 0.39 & 0.80 \\
\midrule
\multicolumn{9}{l}{\textbf{Llama-Generated Content Based on PolitiFact}} \\
\midrule
Sheep's Clothing (Fake) & 7001 & 16.03 & 407.18 & 25.41 & 138.07 & 4.61 & -0.05 & 0.85 \\
Content Manipulation (Fake) & 2445 & 32.49 & 774.30 & 23.83 & 122.90 & 4.37 & 0.35 & 0.85 \\
Information Blending (Fake) & 2000 & 17.28 & 475.45 & 27.52 & 152.29 & 4.70 & 0.05 & 0.87 \\
Narrative Generation (Fake) & 7001 & 12.65 & 347.08 & 27.45 & 151.36 & 4.68 & 0.44 & 0.77  \\
Writing Enhancement (Legitimate) & 2445 & 18.77 & 499.42 & 26.61 & 147.01 & 4.70 & 0.28 & 0.82 \\
News Summarization (Legitimate) & 2000 & 18.64 & 539.02 & 28.92 & 158.63 & 4.65 & 0.26 & 0.85 \\
\bottomrule
\end{tabular}
}
\label{tab:statistics_news}
\end{table}

\subsection{Dataset Consolidation}
Finally, to create a fake news dataset for studying and detecting both human- and machine-generated fake news, we consolidate our LLM-generated data with human-authored fake news from FakeNewsNet~\citep{shu2020fakenewsnet} into a single data repository. The complete dataset is available in our \href{https://osf.io/x7wrs/overview?view_only=cff5842af7b54b168d0479df985a6127}{Online Data Repository}.

\section{Experiments}\label{experiment}
In this section, we present a series of experiments to demonstrate the theoretical and practical value of our MegaFake dataset.

\subsection{Implementation Details}
We employ six LLMs with varying parameter scales, including Qwen \citep{bai2023qwen}, Llama \citep{touvron2023llama}, ChatGLM \citep{glm2024chatglm}, GPT-4 \citep{achiam2023gpt}, Mistral \citep{jiang2023mistral}, and Baichuan \citep{yang2023baichuan}. We conduct zero-shot experiments without fine-tuning on all six LLMs. Besides, we fine-tune several models, including Qwen1.5-7B, Llama3.1-8B, and ChatGLM3-6B, by randomly selecting 32,000 samples from the MegaFake dataset for fine-tuning and 8,000 samples for testing. We apply the LoRA \citep{hu2021lora,ma2026dearfinegrainedvlmadaptation} technique for fine-tuning. For classification, we use the following prompt specified by prior studies \citep{su2024adapting, chen2024llmgenerated}: \texttt{Identify whether the news is legitimate or fake in one word: \{news\}.} All training is conducted on a single NVIDIA A100 GPU.

We also fine-tune several state-of-the-art pre-trained deep learning models for fake news detection. These models include BERT-tiny \citep{jiao-etal-2020-tinybert}, DeBERTa \citep{he2020deberta}, RoBERTa \citep{liu2019roberta}, ELECTRA \citep{clark2020electra}, ALBERT \citep{lan2019albert}, and CT-BERT \citep{muller2023covid}. For model training and testing, we apply 5-fold cross-validation to ensure robustness and minimize variance in our evaluation results. For key training parameters, we use a learning rate of 2e-5, maintain a batch size of 8 for both training and evaluation stages, and train for at least 10 epochs. For optimization, we employ the AdamW optimizer with a weight decay of 0.01. All training is conducted on a single NVIDIA RTX 3090 GPU.

\subsection{Performance in Fake News Detection}
The results are presented in \textbf{Table \ref{tab:2.1(1)}} and \textbf{Table \ref{tab:2.1(2)}}. Overall, our experiments show that deep learning models generally outperform LLMs in fake news detection. However, the performance of LLMs on both datasets improves substantially with fine-tuning. This finding is consistent with previous studies \citep{huang2024fakegpt, jiang2024disinformation, hu2024bad}, which suggest that while LLMs excel at analysis and inference, they are less effective for classification tasks requiring domain-specific knowledge. Notably, the lower performance of Mistral-7B stems primarily from its inherent safety guardrails, which caused the model to consistently refuse to classify news articles as legitimate or fake, resulting in a high failure rate for this specific task.

\begin{table}[!ht]
\centering
\scriptsize
\caption{Model performance on GLM-generated MegaFake.}
\begin{tabularx}{\textwidth}{l c *{3}{>{\centering\arraybackslash}X} *{3}{>{\centering\arraybackslash}X}}
\toprule[1.2pt]
\multirow{2}{*}{Model} & \multirow{2}{*}{Accuracy} & \multicolumn{3}{c}{Legitimate$^1$} & \multicolumn{3}{c}{Fake$^1$} \\
\cmidrule(lr){3-5} \cmidrule(lr){6-8}
& & Precision & Recall & F1-Score & Precision & Recall & F1-Score \\
\midrule
\multicolumn{8}{l}{\textbf{Results Based on LLM Performance}} \\
\midrule
\textsc{Qwen1.5-7B} & $0.6478$ & $0.7246$ & $0.8572$ & $0.7854$ & $0.1044$ & $0.0451$ & $0.0630$ \\
\textsc{Qwen1.5-7B\textsubscript{LoRA}} & $0.9113$ & $0.9492$ & $0.9303$ & $0.9397$ & $0.8104$ & $0.8567$ & $0.8329$ \\
\textsc{Qwen1.5-72B} & $0.6090$ & $0.7078$ & $0.8061$ & $0.7538$ & $0.0699$ & $0.0415$ & $0.0520$ \\
\textsc{LLaMA3.1-8B} & $0.5564$ & $0.6969$ & $0.7181$ & $0.7073$ & $0.1126$ & $0.0910$ & $0.1007$ \\
\textsc{LLaMA3.1-8B\textsubscript{LoRA}} & $0.9160$ & $0.9530$ & $0.9328$ & $0.9428$ & $0.8179$ & $0.8676$ & $0.8420$ \\
\textsc{LLaMA2-70B} & $0.5132$ & $0.6752$ & $0.6528$ & $0.6638$ & $0.1129$ & $0.1233$ & $0.1179$ \\
\textsc{LLaMA3-70B} & $0.5445$ & $0.6694$ & $0.7310$ & $0.6988$ & $0.0760$ & $0.0578$ & $0.0656$ \\
\textsc{ChatGLM3-6B} & $0.4908$ & $0.6974$ & $0.6072$ & $0.6492$ & $0.1564$ & $0.1558$ & $0.1561$ \\
\textsc{ChatGLM3-6B\textsubscript{LoRA}} & $0.9085$ & $0.9471$ & $0.9288$ & $0.9379$ & $0.8061$ & $0.8502$ & $0.8276$ \\
\cdashline{1-8}
\textsc{Mistral-7B} & $0.0143$ & $0.5259$ & $0.0081$ & $0.0159$ & $0.3032$ & $0.0323$ & $0.0584$ \\
\textsc{Mistral-7B-Instruct} & $0.6140$ & $0.7140$ & $0.8042$ & $0.7564$ & $0.1109$ & $0.0665$ & $0.0831$ \\
\textsc{Mistral-8×7B} & $0.6752$ & $0.7199$ & $0.9138$ & $0.8053$ & $0.0508$ & $0.0128$ & $0.0205$ \\
\textsc{Baichuan-7B} & $0.3563$ & $0.7106$ & $0.4748$ & $0.5693$ & $0.0877$ & $0.0151$ & $0.0257$ \\
\textsc{GPT-4o} & $0.4215$ & $0.6381$ & $0.4808$ & $0.5458$ & $0.1495$ & $0.2474$ & $0.1864$ \\
\midrule
\multicolumn{8}{l}{\textbf{Results Based on Deep Learning Model Performance}$^1$ $^2$} \\
\midrule
\textsc{DeBERTa-Base} &
$\underset{\text{\tiny(0.0025)}}{0.9086}$ &
$\underset{\text{\tiny(0.0037)}}{0.9357}$ &
$\underset{\text{\tiny(0.0048)}}{0.9415}$ &
$\underset{\text{\tiny(0.0017)}}{0.9386}$ &
$\underset{\text{\tiny(0.0104)}}{0.8288}$ &
$\underset{\text{\tiny(0.0120)}}{0.8138}$ &
$\underset{\text{\tiny(0.0049)}}{0.8211}$ \\
\textsc{BERT-Base} &
$\underset{\text{\tiny(0.0023)}}{0.9008}$ &
$\underset{\text{\tiny(0.0047)}}{0.9295}$ &
$\underset{\text{\tiny(0.0056)}}{0.9374}$ &
$\underset{\text{\tiny(0.0016)}}{0.9334}$ &
$\underset{\text{\tiny(0.0113)}}{0.8155}$ &
$\underset{\text{\tiny(0.0159)}}{0.7953}$ &
$\underset{\text{\tiny(0.0053)}}{0.8051}$ \\
\textsc{ELECTRA-Base} &
$\underset{\text{\tiny(0.0033)}}{0.8936}$ &
$\underset{\text{\tiny(0.0051)}}{0.9212}$ &
$\underset{\text{\tiny(0.0022)}}{0.9369}$ &
$\underset{\text{\tiny(0.0020)}}{0.9290}$ &
$\underset{\text{\tiny(0.0039)}}{0.8089}$ &
$\underset{\text{\tiny(0.0166)}}{0.7692}$ &
$\underset{\text{\tiny(0.0087)}}{0.7885}$ \\
\textsc{ALBERT-Base} &
$\underset{\text{\tiny(0.0040)}}{0.9010}$ &
$\underset{\text{\tiny(0.0019)}}{0.9337}$ &
$\underset{\text{\tiny(0.0058)}}{0.9328}$ &
$\underset{\text{\tiny(0.0029)}}{0.9333}$ &
$\underset{\text{\tiny(0.0128)}}{0.8074}$ &
$\underset{\text{\tiny(0.0063)}}{0.8094}$ &
$\underset{\text{\tiny(0.0064)}}{0.8084}$ \\
\textsc{CT-BERT} &
$\underset{\text{\tiny(0.0487)}}{0.7819}$ &
$\underset{\text{\tiny(0.0707)}}{0.7999}$ &
$\underset{\text{\tiny(0.0480)}}{0.9608}$ &
$\underset{\text{\tiny(0.0207)}}{0.8689}$ &
$\underset{\text{\tiny(0.3200)}}{0.3812}$ &
$\underset{\text{\tiny(0.3270)}}{0.2671}$ &
$\underset{\text{\tiny(0.3354)}}{0.2741}$ \\
\textsc{RoBERTa-Base} &
$\underset{\text{\tiny(0.0028)}}{0.9060}$ &
$\underset{\text{\tiny(0.0060)}}{0.9297}$ &
$\underset{\text{\tiny(0.0060)}}{0.9448}$ &
$\underset{\text{\tiny(0.0018)}}{0.9372}$ &
$\underset{\text{\tiny(0.0122)}}{0.8338}$ &
$\underset{\text{\tiny(0.0198)}}{0.7942}$ &
$\underset{\text{\tiny(0.0072)}}{0.8132}$ \\
\bottomrule[1.2pt]
\multicolumn{8}{l}{$^1$ Both the fake and legitimate classes contain content generated by humans and machines.} \\
\multicolumn{8}{l}{$^2$ For deep learning models, results are reported as the average across 5-fold cross-validation, with standard deviation shown in parentheses.}
\end{tabularx}
\label{tab:2.1(1)}
\end{table}

\begin{table}[!ht]
\centering
\scriptsize
\caption{Model performance on Llama-generated MegaFake.}
\begin{tabularx}{\textwidth}{l c *{3}{>{\centering\arraybackslash}X} *{3}{>{\centering\arraybackslash}X}}
\toprule[1.2pt]
\multirow{2}{*}{Model} & \multirow{2}{*}{Accuracy} & \multicolumn{3}{c}{Legitimate$^1$} & \multicolumn{3}{c}{Fake$^1$} \\
\cmidrule(lr){3-5} \cmidrule(lr){6-8}
 &  & Precision & Recall & F1-Score & Precision & Recall & F1-Score \\
\midrule
\multicolumn{8}{l}{\textbf{Results Based on LLM Performance}} \\
\midrule
\textsc{Qwen1.5-7B} & $0.6906$ & $0.7717$ & $0.8506$ & $0.8093$ & $0.2416$ & $0.1554$ & $0.1892$ \\
\textsc{Qwen1.5-7B\textsubscript{LoRA}} & $0.9587$ & $0.9755$ & $0.9708$ & $0.9731$ & $0.9040$ & $0.9184$ & $0.9111$ \\
\textsc{Qwen1.5-72B} & $0.6537$ & $0.7582$ & $0.8080$ & $0.7823$ & $0.1771$ & $0.1374$ & $0.1547$ \\
\textsc{LLaMA3.1-8B} & $0.5882$ & $0.7383$ & $0.7316$ & $0.7349$ & $0.1192$ & $0.1087$ & $0.1137$ \\
\textsc{LLaMA3.1-8B\textsubscript{LoRA}} & $0.9583$ & $0.9787$ & $0.9668$ & $0.9727$ & $0.8933$ & $0.9297$ & $0.9111$ \\
\textsc{LLaMA2-70B} & $0.5538$ & $0.7231$ & $0.6766$ & $0.6991$ & $0.1258$ & $0.1522$ & $0.1377$ \\
\textsc{LLaMA3-70B} & $0.6299$ & $0.7338$ & $0.8110$ & $0.7704$ & $0.0570$ & $0.0374$ & $0.0451$ \\
\textsc{ChatGLM3-6B} & $0.5202$ & $0.7533$ & $0.5973$ & $0.6663$ & $0.2030$ & $0.2622$ & $0.2289$ \\
\textsc{ChatGLM3-6B\textsubscript{LoRA}} & $0.9559$ & $0.9736$ & $0.9690$ & $0.9713$ & $0.8979$ & $0.9120$ & $0.9049$ \\
\cdashline{1-8}
\textsc{Mistral-7B} & $0.0244$ & $0.6575$ & $0.0118$ & $0.0232$ & $0.2939$ & $0.0663$ & $0.1082$ \\
\textsc{Mistral-7B-Instruct} & $0.6416$ & $0.7447$ & $0.8141$ & $0.7778$ & $0.0971$ & $0.0647$ & $0.0776$ \\
\textsc{Mistral-8×7B} & $0.7015$ & $0.7504$ & $0.9136$ & $0.8240$ & $0.0366$ & $0.0107$ & $0.0165$ \\
\textsc{Baichuan-7B} & $0.2745$ & $0.7368$ & $0.3526$ & $0.4769$ & $0.0736$ & $0.0135$ & $0.0228$ \\
\textsc{GPT-4o} & $0.4163$ & $0.6309$ & $0.4808$ & $0.5458$ & $0.1479$ & $0.2426$ & $0.1838$ \\
\midrule
\multicolumn{8}{l}{\textbf{Results Based on Deep Learning Model Performance}$^1$ $^2$} \\
\midrule
\textsc{DeBERTa-Base}  & $\underset{\text{\tiny(0.0014)}}{0.9494}$ & $\underset{\text{\tiny(0.0026)}}{0.9610}$ & $\underset{\text{\tiny(0.0024)}}{0.9737}$ & $\underset{\text{\tiny(0.0009)}}{0.9673}$ & $\underset{\text{\tiny(0.0069)}}{0.9082}$ & $\underset{\text{\tiny(0.0093)}}{0.8678}$ & $\underset{\text{\tiny(0.0035)}}{0.8875}$ \\
\textsc{BERT-Base}     & $\underset{\text{\tiny(0.0014)}}{0.9454}$ & $\underset{\text{\tiny(0.0029)}}{0.9594}$ & $\underset{\text{\tiny(0.0042)}}{0.9701}$ & $\underset{\text{\tiny(0.0010)}}{0.9647}$ & $\underset{\text{\tiny(0.0116)}}{0.8966}$ & $\underset{\text{\tiny(0.0108)}}{0.8627}$ & $\underset{\text{\tiny(0.0027)}}{0.8792}$ \\
\textsc{ELECTRA-Base}  & $\underset{\text{\tiny(0.0005)}}{0.9441}$ & $\underset{\text{\tiny(0.0046)}}{0.9578}$ & $\underset{\text{\tiny(0.0052)}}{0.9702}$ & $\underset{\text{\tiny(0.0004)}}{0.9639}$ & $\underset{\text{\tiny(0.0142)}}{0.8963}$ & $\underset{\text{\tiny(0.0171)}}{0.8570}$ & $\underset{\text{\tiny(0.0023)}}{0.8759}$ \\
\textsc{ALBERT-Base}   & $\underset{\text{\tiny(0.0017)}}{0.9430}$ & $\underset{\text{\tiny(0.0020)}}{0.9608}$ & $\underset{\text{\tiny(0.0009)}}{0.9653}$ & $\underset{\text{\tiny(0.0011)}}{0.9631}$ & $\underset{\text{\tiny(0.0026)}}{0.8820}$ & $\underset{\text{\tiny(0.0071)}}{0.8684}$ & $\underset{\text{\tiny(0.0041)}}{0.8752}$ \\
\textsc{CT-BERT}       & $\underset{\text{\tiny(0.0770)}}{0.8489}$ & $\underset{\text{\tiny(0.0871)}}{0.8547}$ & $\underset{\text{\tiny(0.0182)}}{0.9849}$ & $\underset{\text{\tiny(0.0414)}}{0.9120}$ & $\underset{\text{\tiny(0.1399)}}{0.8207}$ & $\underset{\text{\tiny(0.3933)}}{0.3939}$ & $\underset{\text{\tiny(0.3872)}}{0.4260}$ \\
\textsc{RoBERTa-Base}  & $\underset{\text{\tiny(0.0017)}}{0.9483}$ & $\underset{\text{\tiny(0.0021)}}{0.9607}$ & $\underset{\text{\tiny(0.0028)}}{0.9726}$ & $\underset{\text{\tiny(0.0011)}}{0.9666}$ & $\underset{\text{\tiny(0.0081)}}{0.9046}$ & $\underset{\text{\tiny(0.0075)}}{0.8668}$ & $\underset{\text{\tiny(0.0035)}}{0.8853}$ \\

\bottomrule[1.2pt]
\multicolumn{8}{l}{$^1$ Both the fake and legitimate classes contain content generated by humans and machines.} \\
\multicolumn{8}{l}{$^2$ For deep learning models, results are reported as the average across 5-fold cross-validation, with standard deviation shown in parentheses.}

\end{tabularx}
\label{tab:2.1(2)}
\end{table}

Specifically, for the GLM-generated MegaFake dataset (see \textbf{Table \ref{tab:2.1(1)}}), the results clearly show that deep learning models outperform LLMs without fine-tuning across all major evaluation metrics, including accuracy, F1-score, and class-level precision and recall. Among deep learning models, DeBERTa-Base, BERT-Base, ALBERT-Base, and RoBERTa-Base consistently achieve the highest scores, with overall accuracy exceeding 0.90. Some fine-tuned LLMs, such as Llama3.1-8B, are able to slightly surpass this standard, and they demonstrate substantial improvements compared to LLMs without fine-tuning. LoRA fine-tuning consistently enhances LLM performance, as evidenced by both Qwen1.5-7B and ChatGLM3-6B, where the LoRA variants significantly outperform their base counterparts. These findings highlight the importance of domain-specific adaptation for LLMs in fake news detection.

Moreover, most deep learning models and all fine-tuned LLMs perform better at detecting legitimate news than fake news. For example, DeBERTa-Base achieves an F1-score of 0.9386 for legitimate news but only 0.8211 for fake news. Similarly, the strongest LLMs tend to have high F1-scores for legitimate news (e.g., 0.9428 for Llama3.1-8B), but their ability to identify fake news remains relatively lower (F1-score of 0.8420 for fake news). This disparity suggests that while LLMs excel at recognizing legitimate content, they are less reliable when it comes to detecting more subtle or sophisticated fake news.

These findings are consistent with the results on the Llama-generated MegaFake dataset, as shown in \textbf{Table \ref{tab:2.1(2)}}. Overall, the metrics closely mirror those observed on the GLM-generated dataset, reinforcing the reliability and robustness of our conclusions. Specifically, deep learning models continue to outperform LLMs without fine-tuning by a considerable margin across accuracy, F1-score, and class-specific precision and recall. Among LLMs, LoRA fine-tuning again proves critical for boosting performance. For example, Qwen1.5-7B, Llama3.1-8B, and ChatGLM3-6B show substantial improvements over their base versions, achieving accuracy and F1-scores that surpass those of the best deep learning models. Notably, these fine-tuned LLMs maintain a strong balance between legitimate and fake news detection, with F1-scores above 0.90 for both classes in the best cases.

\subsection{Performance in Machine-Generated Content Detection}\label{f.2}

\subsubsection{Experiments on Machine-Generated Legitimate News}
We first conduct experiments on legitimate news within the GLM-generated MegaFake dataset. \textbf{Table \ref{tab:2.1(3)C}} shows that deep learning models perform comparably to LLMs in distinguishing between \say{news summarization} and \say{writing enhancement} legitimate news types. Specifically, RoBERTa-Base achieves the best performance among the deep learning models, with relatively higher accuracy and consistently high precision, recall, and F1-scores across both news categories. Among the LLMs, Mistral-7B achieves the best performance in this task, with an accuracy of 0.7731.

\begin{table}[!ht]
\centering
\scriptsize
\caption{Model performance on classifying machine-generated legitimate news in GLM-generated MegaFake.}
\begin{tabularx}{\textwidth}{l c *{3}{>{\centering\arraybackslash}X} *{3}{>{\centering\arraybackslash}X}}
\toprule[1.2pt]
\multirow{2}{*}{Model} & \multirow{2}{*}{Accuracy} &
\multicolumn{3}{c}{News Summarization} &
\multicolumn{3}{c}{Writing Enhancement} \\
\cmidrule(lr){3-5} \cmidrule(lr){6-8}
& & Precision & Recall & F1-Score & Precision & Recall & F1-Score \\
\midrule
\multicolumn{8}{l}{\textbf{Results Based on LLM Performance}} \\
\midrule
\textsc{Baichuan-7B}   & $0.4478$ & $0.3521$ & $0.7335$ & $0.4758$ & $0.6836$ & $0.3858$ & $0.4932$ \\
\textsc{LLaMA3.1-8B}   & $0.4696$ & $0.3795$ & $0.6525$ & $0.4799$ & $0.6405$ & $0.6049$ & $0.6222$ \\
\textsc{Mistral-7B-Instruct} & $0.7731$ & $0.4030$ & $0.0092$ & $0.0181$ & $0.7773$ & $0.9936$ & $0.8722$ \\
\textsc{Qwen1.5-7B}   & $0.5754$ & $0.3507$ & $0.3912$ & $0.3699$ & $0.7107$ & $0.7048$ & $0.7077$ \\
\textsc{ChatGLM3-6B}  & $0.6194$ & $0.2909$ & $0.0940$ & $0.1421$ & $0.6445$ & $0.9975$ & $0.7831$ \\
\midrule
\multicolumn{8}{l}{\textbf{Results Based on Deep Learning Model Performance}$^1$} \\
\midrule
\textsc{DeBERTa-Base}  & $\underset{\text{\tiny(0.0029)}}{0.5014}$ & $\underset{\text{\tiny(0.1056)}}{0.3028}$ & $\underset{\text{\tiny(0.0029)}}{0.5014}$ & $\underset{\text{\tiny(0.0124)}}{0.6605}$ & $\underset{\text{\tiny(0.1849)}}{0.4392}$ & $\underset{\text{\tiny(0.0152)}}{0.4906}$ & $\underset{\text{\tiny(0.0981)}}{0.6052}$ \\
\textsc{BERT-Base}     & $\underset{\text{\tiny(0.0070)}}{0.4929}$ & $\underset{\text{\tiny(0.0458)}}{0.4939}$ & $\underset{\text{\tiny(0.0070)}}{0.4929}$ & $\underset{\text{\tiny(0.0911)}}{0.1160}$ & $\underset{\text{\tiny(0.1215)}}{0.5301}$ & $\underset{\text{\tiny(0.0188)}}{0.5090}$ & $\underset{\text{\tiny(0.1663)}}{0.0863}$ \\
\textsc{ELECTRA-Base}  & $\underset{\text{\tiny(0.0399)}}{0.5134}$ & $\underset{\text{\tiny(0.0481)}}{0.5202}$ & $\underset{\text{\tiny(0.0399)}}{0.5134}$ & $\underset{\text{\tiny(0.0840)}}{0.5415}$ & $\underset{\text{\tiny(0.0025)}}{0.5222}$ & $\underset{\text{\tiny(0.0034)}}{0.5209}$ & $\underset{\text{\tiny(0.0750)}}{0.4902}$ \\
\textsc{CT-BERT} & $\underset{\text{\tiny(0.0144)}}{0.5294}$ & $\underset{\text{\tiny(0.0546)}}{0.5683}$ & $\underset{\text{\tiny(0.1837)}}{0.4187}$ & $\underset{\text{\tiny(0.1059)}}{0.4508}$ & $\underset{\text{\tiny(0.0198)}}{0.5290}$ & $\underset{\text{\tiny(0.2969)}}{0.6469}$ & $\underset{\text{\tiny(0.1410)}}{0.5402}$ \\
\textsc{ALBERT-Base}   & $\underset{\text{\tiny(0.0018)}}{0.5009}$ & $\underset{\text{\tiny(0.1172)}}{0.3085}$ & $\underset{\text{\tiny(0.0018)}}{0.5009}$ & $\underset{\text{\tiny(0.2664)}}{0.5327}$ & $\underset{\text{\tiny(0.1275)}}{0.3542}$ & $\underset{\text{\tiny(0.0010)}}{0.5008}$ & $\underset{\text{\tiny(0.2506)}}{0.1756}$ \\
\textsc{RoBERTa-Base} & $\underset{\text{\tiny(0.0155)}}{0.6145}$ & $\underset{\text{\tiny(0.0493)}}{0.6118}$ & $\underset{\text{\tiny(0.1705)}}{0.6913}$ & $\underset{\text{\tiny(0.0655)}}{0.6313}$ & $\underset{\text{\tiny(0.0510)}}{0.6347}$ & $\underset{\text{\tiny(0.1702)}}{0.6038}$ & $\underset{\text{\tiny(0.0922)}}{0.5964}$ \\
\bottomrule[1.2pt]
\multicolumn{8}{l}{$^1$ For deep learning models, results are reported as the average across 5-fold cross-validation, with standard deviation shown in parentheses.}
\end{tabularx}
\label{tab:2.1(3)C}
\end{table}

The results presented in \textbf{Table \ref{tab:2.1(3)L}}, which are based on the Llama-generated MegaFake dataset for classifying legitimate news types, are highly consistent with the findings from the GLM-generated dataset. Across both datasets, deep learning models and LLMs perform similarly in distinguishing between \say{news summarization} and \say{writing enhancement} legitimate news types. ELECTRA-Base achieves a high accuracy exceeding 0.67, while Mistral-7B again attains the highest accuracy among all detection models, with a score of 0.7474. Overall, these results indicate that this new task poses significant challenges for both LLMs and deep learning models. Therefore, we anticipate that further research will be needed to develop more effective models for identifying machine-generated legitimate news.

\begin{table}[!ht]
\centering
\scriptsize
\caption{Model performance on classifying machine-generated legitimate news in Llama-generated MegaFake.}
\begin{tabularx}{\textwidth}{l c *{3}{>{\centering\arraybackslash}X} *{3}{>{\centering\arraybackslash}X}}
\toprule[1.2pt]
\multirow{2}{*}{Model} & \multirow{2}{*}{Accuracy} &
\multicolumn{3}{c}{News Summarization} &
\multicolumn{3}{c}{Writing Enhancement} \\
\cmidrule(lr){3-5} \cmidrule(lr){6-8}
& & Precision & Recall & F1-Score & Precision & Recall & F1-Score \\
\midrule
\multicolumn{8}{l}{\textbf{Results Based on LLM Performance}} \\
\midrule
\textsc{Baichuan-7B} & $0.4293$ & $0.2625$ & $0.9146$ & $0.4080$ & $0.8083$ & $0.5460$ & $0.6518$ \\
\textsc{LLaMA3.1-8B} & $0.4624$ & $0.3019$ & $0.7986$ & $0.4381$ & $0.7308$ & $0.5749$ & $0.6435$ \\
\textsc{Mistral-7B-Instruct} & $0.7474$ & $0.0828$ & $0.0007$ & $0.0013$ & $0.1034$ & $0.0005$ & $0.0011$ \\
\textsc{Qwen1.5-7B} & $0.6414$ & $0.2328$ & $0.2748$ & $0.2521$ & $0.7806$ & $0.8352$ & $0.8070$ \\
\textsc{ChatGLM3-6B} & $0.7118$ & $0.3159$ & $0.1954$ & $0.2414$ & $0.7419$ & $0.9983$ & $0.8512$ \\
\midrule
\multicolumn{8}{l}{\textbf{Results Based on Deep Learning Model Performance}$^1$} \\
\midrule
\textsc{DeBERTa-Base}  & $\underset{\text{\tiny(0.0343)}}{0.5170}$ & $\underset{\text{\tiny(0.1804)}}{0.3864}$ & $\underset{\text{\tiny(0.0343)}}{0.5170}$ & $\underset{\text{\tiny(0.0132)}}{0.6727}$ & $\underset{\text{\tiny(0.0590)}}{0.5285}$ & $\underset{\text{\tiny(0.1713)}}{0.3354}$ & $\underset{\text{\tiny(0.0454)}}{0.6431}$ \\
\textsc{BERT-Base}     & $\underset{\text{\tiny(0.1297)}}{0.6446}$ & $\underset{\text{\tiny(0.2816)}}{0.5937}$ & $\underset{\text{\tiny(0.1297)}}{0.6446}$ & $\underset{\text{\tiny(0.3289)}}{0.3865}$ & $\underset{\text{\tiny(0.0077)}}{0.5051}$ & $\underset{\text{\tiny(0.0697)}}{0.6730}$ & $\underset{\text{\tiny(0.0360)}}{0.0244}$ \\
\textsc{ELECTRA-Base}  & $\underset{\text{\tiny(0.0694)}}{0.6734}$ & $\underset{\text{\tiny(0.0681)}}{0.7061}$ & $\underset{\text{\tiny(0.0694)}}{0.6734}$ & $\underset{\text{\tiny(0.1486)}}{0.6399}$ & $\underset{\text{\tiny(0.1000)}}{0.3880}$ & $\underset{\text{\tiny(0.1424)}}{0.3972}$ & $\underset{\text{\tiny(0.0552)}}{0.3647}$ \\
\textsc{CT-BERT} & $\underset{\text{\tiny(0.0434)}}{0.5289}$ & $\underset{\text{\tiny(0.1112)}}{0.8846}$ & $\underset{\text{\tiny(0.1331)}}{0.1128}$ & $\underset{\text{\tiny(0.1945)}}{0.1679}$ & $\underset{\text{\tiny(0.3891)}}{0.1945}$ & $\underset{\text{\tiny(0.0712)}}{0.0356}$ & $\underset{\text{\tiny(0.1204)}}{0.0602}$ \\
\textsc{ALBERT-Base}   & $\underset{\text{\tiny(0.0027)}}{0.5026}$ & $\underset{\text{\tiny(0.2199)}}{0.5159}$ & $\underset{\text{\tiny(0.0027)}}{0.5026}$ & $\underset{\text{\tiny(0.0008)}}{0.6675}$ & $\underset{\text{\tiny(0.0613)}}{0.5304}$ & $\underset{\text{\tiny(0.1854)}}{0.3426}$ & $\underset{\text{\tiny(0.0238)}}{0.6783}$ \\
\textsc{RoBERTa-Base} & $\underset{\text{\tiny(0.1352)}}{0.6124}$ & $\underset{\text{\tiny(0.1559)}}{0.6931}$ & $\underset{\text{\tiny(0.1376)}}{0.8918}$ & $\underset{\text{\tiny(0.0713)}}{0.7535}$ & $\underset{\text{\tiny(0.0364)}}{0.5182}$ & $\underset{\text{\tiny(0.0000)}}{1.0000}$ & $\underset{\text{\tiny(0.0305)}}{0.6819}$ \\

\bottomrule[1.2pt]
\multicolumn{8}{l}{$^1$ For deep learning models, results are reported as the average across 5-fold cross-validation, with standard deviation shown in parentheses.}
\end{tabularx}
\label{tab:2.1(3)L}
\end{table}

\subsubsection{Experiments on Machine-Generated Fake News}
Next, we examine the performance of various deep learning models and LLMs in classifying machine-generated fake news types. We first conduct experiments on machine-generated fake news within the GLM-generated MegaFake dataset, with results presented in \textbf{Table \ref{tab:2.1(4)C}}. LLMs such as Baichuan-7B and ChatGLM3-6B perform poorly in classifying machine-generated fake news, particularly for the \say{information blending} news type, where precision drops below 0.1. This finding suggests that LLMs exhibit limited generalization capabilities for these tasks. In contrast, deep learning models such as CT-BERT and RoBERTa-Base achieve relatively better performance across all categories, especially for the \say{content manipulation} and \say{information blending} fake news types, as evidenced by higher precision, recall, and F1-scores. Nonetheless, all detection models generally struggle to accurately identify machine-generated fake news.

\setlength{\tabcolsep}{3pt}
\begin{table}[!ht]
\centering
\scriptsize
\renewcommand{\arraystretch}{1.5}
\caption{Model performance on classifying machine-generated fake news in GLM-generated MegaFake.}
\begin{adjustbox}{max width=\linewidth}
\begin{tabular}{lccccccccccccc}
\toprule[1.2pt]
\multirow{2}{*}{Model} & \multirow{2}{*}{Accuracy} &
\multicolumn{3}{c}{Content Manipulation} &
\multicolumn{3}{c}{Information Blending} &
\multicolumn{3}{c}{Narrative Generation} &
\multicolumn{3}{c}{Sheep's Clothing} \\
\cmidrule(lr){3-5} \cmidrule(lr){6-8} \cmidrule(lr){9-11} \cmidrule(lr){12-14}
& & Precision & Recall & F1-Score & Precision & Recall & F1-Score & Precision & Recall & F1-Score & Precision & Recall & F1-Score \\
\midrule
\multicolumn{14}{l}{\textbf{Results Based on LLM Performance}} \\
\midrule
\textsc{Baichuan-7B} & $0.3806$ & $0.2389$ & $0.6866$ & $0.3545$ & $0.0631$ & $0.6208$ & $0.1145$ & $0.3527$ & $0.9267$ & $0.5109$ & $0.3764$ & $0.7514$ & $0.5015$ \\
\textsc{LLaMA3.1-8B} & $0.4906$ & $0.2295$ & $0.2944$ & $0.2579$ & $0.0945$ & $0.4758$ & $0.1577$ & $0.3376$ & $0.6459$ & $0.4434$ & $0.3484$ & $0.8021$ & $0.4858$ \\
\textsc{Mistral-7B-Instruct} & $0.7465$ & $0.4030$ & $0.0092$ & $0.0181$ & $0.0127$ & $0.0006$ & $0.0011$ & $0.0181$ & $0.0003$ & $0.0006$ & $0.4145$ & $0.0035$ & $0.0070$ \\
\textsc{Qwen1.5-7B} & $0.5188$ & $0.2898$ & $0.3819$ & $0.3296$ & $0.0521$ & $0.4043$ & $0.0922$ & $0.2815$ & $0.0178$ & $0.0336$ & $0.3532$ & $0.9572$ & $0.5160$ \\
\textsc{ChatGLM3-6B} & $0.3052$ & $0.2352$ & $0.9387$ & $0.3761$ & $0.0690$ & $0.8346$ & $0.1274$ & $0.3660$ & $0.8890$ & $0.5185$ & $0.3525$ & $0.9908$ & $0.5200$ \\
\midrule
\multicolumn{14}{l}{\textbf{Results Based on Deep Learning Model Performance}$^1$} \\
\midrule
\textsc{Deberta-Base} & $\underset{\text{\tiny(0.0346)}}{0.5107}$ & 
$\underset{\text{\tiny(0.0767)}}{0.5613}$ & 
$\underset{\text{\tiny(0.2472)}}{0.7911}$ & 
$\underset{\text{\tiny(0.0568)}}{0.6226}$ & 
$\underset{\text{\tiny(0.0028)}}{0.5015}$ & 
$\underset{\text{\tiny(0.0324)}}{0.9838}$ & 
$\underset{\text{\tiny(0.0053)}}{0.6641}$ & 
$\underset{\text{\tiny(0.0119)}}{0.4940}$ & 
$\underset{\text{\tiny(0.0974)}}{0.9513}$ & 
$\underset{\text{\tiny(0.0347)}}{0.6493}$ & 
$\underset{\text{\tiny(0.0026)}}{0.5013}$ & 
$\underset{\text{\tiny(0.0260)}}{0.9870}$ & 
$\underset{\text{\tiny(0.0038)}}{0.6647}$ \\

\textsc{Bert-Base} & $\underset{\text{\tiny(0.0702)}}{0.5414}$ & 
$\underset{\text{\tiny(0.1791)}}{0.8135}$ & 
$\underset{\text{\tiny(0.1628)}}{0.3152}$ & 
$\underset{\text{\tiny(0.2165)}}{0.4324}$ & 
$\underset{\text{\tiny(0.2009)}}{0.3319}$ & 
$\underset{\text{\tiny(0.3099)}}{0.2540}$ & 
$\underset{\text{\tiny(0.2830)}}{0.2344}$ & 
$\underset{\text{\tiny(0.1764)}}{0.4297}$ & 
$\underset{\text{\tiny(0.1262)}}{0.0664}$ & 
$\underset{\text{\tiny(0.1393)}}{0.0761}$ & 
$\underset{\text{\tiny(0.1933)}}{0.5119}$ & 
$\underset{\text{\tiny(0.2149)}}{0.1878}$ & 
$\underset{\text{\tiny(0.2263)}}{0.2242}$ \\

\textsc{Electra-Base} & $\underset{\text{\tiny(0.0539)}}{0.5035}$ & 
$\underset{\text{\tiny(0.0776)}}{0.4239}$ & 
$\underset{\text{\tiny(0.1759)}}{0.4097}$ & 
$\underset{\text{\tiny(0.1321)}}{0.4080}$ & 
$\underset{\text{\tiny(0.0159)}}{0.5236}$ & 
$\underset{\text{\tiny(0.1755)}}{0.6483}$ & 
$\underset{\text{\tiny(0.0686)}}{0.5703}$ & 
$\underset{\text{\tiny(0.0341)}}{0.5415}$ & 
$\underset{\text{\tiny(0.0694)}}{0.4879}$ & 
$\underset{\text{\tiny(0.0327)}}{0.5078}$ & 
$\underset{\text{\tiny(0.0266)}}{0.5027}$ & 
$\underset{\text{\tiny(0.2139)}}{0.6176}$ & 
$\underset{\text{\tiny(0.0978)}}{0.5388}$ \\

\textsc{Ct-Bert} & $\underset{\text{\tiny(0.0484)}}{0.5556}$ & 
$\underset{\text{\tiny(0.1031)}}{0.6728}$ & 
$\underset{\text{\tiny(0.2854)}}{0.4827}$ & 
$\underset{\text{\tiny(0.2221)}}{0.4900}$ & 
$\underset{\text{\tiny(0.0516)}}{0.5599}$ & 
$\underset{\text{\tiny(0.1934)}}{0.7394}$ & 
$\underset{\text{\tiny(0.0465)}}{0.6180}$ & 
$\underset{\text{\tiny(0.0152)}}{0.5276}$ & 
$\underset{\text{\tiny(0.0728)}}{0.6012}$ & 
$\underset{\text{\tiny(0.0387)}}{0.5605}$ & 
$\underset{\text{\tiny(0.0288)}}{0.5307}$ & 
$\underset{\text{\tiny(0.1676)}}{0.7367}$ & 
$\underset{\text{\tiny(0.0548)}}{0.6037}$ \\

\textsc{Albert-Base} & $\underset{\text{\tiny(0.0310)}}{0.5172}$ & 
$\underset{\text{\tiny(0.2093)}}{0.4146}$ & 
$\underset{\text{\tiny(0.4837)}}{0.6017}$ & 
$\underset{\text{\tiny(0.3173)}}{0.4066}$ & 
$\underset{\text{\tiny(0.0093)}}{0.5049}$ & 
$\underset{\text{\tiny(0.0222)}}{0.9879}$ & 
$\underset{\text{\tiny(0.0104)}}{0.6681}$ & 
$\underset{\text{\tiny(0.1947)}}{0.6131}$ & 
$\underset{\text{\tiny(0.3771)}}{0.5409}$ & 
$\underset{\text{\tiny(0.2576)}}{0.4474}$ & 
$\underset{\text{\tiny(0.0499)}}{0.5329}$ & 
$\underset{\text{\tiny(0.2014)}}{0.8891}$ & 
$\underset{\text{\tiny(0.0497)}}{0.6473}$ \\

\textsc{Roberta-Base} & $\underset{\text{\tiny(0.0659)}}{0.5738}$ & 
$\underset{\text{\tiny(0.0752)}}{0.9277}$ & 
$\underset{\text{\tiny(0.0597)}}{0.3746}$ & 
$\underset{\text{\tiny(0.0395)}}{0.5273}$ & 
$\underset{\text{\tiny(0.2139)}}{0.4249}$ & 
$\underset{\text{\tiny(0.3955)}}{0.7908}$ & 
$\underset{\text{\tiny(0.2768)}}{0.5523}$ & 
$\underset{\text{\tiny(0.2936)}}{0.5441}$ & 
$\underset{\text{\tiny(0.3869)}}{0.4113}$ & 
$\underset{\text{\tiny(0.2658)}}{0.3772}$ & 
$\underset{\text{\tiny(0.2850)}}{0.3427}$ & 
$\underset{\text{\tiny(0.4625)}}{0.4125}$ & 
$\underset{\text{\tiny(0.3107)}}{0.3100}$ \\

\bottomrule[1.2pt]
\multicolumn{14}{l}{$^1$ For deep learning models, results are reported as the average across 5-fold cross-validation, with standard deviation shown in parentheses.}
\end{tabular}
\end{adjustbox}
\label{tab:2.1(4)C}
\end{table}

\textbf{Table \ref{tab:2.1(4)L}} presents the results for the Llama-generated MegaFake dataset, revealing similar patterns and notable disparities between the performance of LLMs and deep learning models. Deep learning models—particularly RoBERTa-Base, tend to outperform LLMs across most metrics, except for Mistral-7B, which achieves the highest accuracy but with highly imbalanced recall and precision. Overall, these results highlight the significant challenges that current detection models face in the machine-generated fake news detection task. Despite recent advances and the increasing scale of deep learning models and LLMs, these models remain inefficient for this specific task, highlighting a critical risk that AI-driven deception and deepfake text pose to the online environment.

\setlength{\tabcolsep}{3pt}
\begin{table}[!ht]
\centering
\scriptsize
\renewcommand{\arraystretch}{1.5}
\caption{Model performance on classifying machine-generated fake news in Llama-generated MegaFake.}
\begin{adjustbox}{max width=\linewidth}
\begin{tabular}{lccccccccccccc}
\toprule[1.2pt]
\multirow{2}{*}{Model} & \multirow{2}{*}{Accuracy} &
\multicolumn{3}{c}{Content Manipulation} &
\multicolumn{3}{c}{Information Blending} &
\multicolumn{3}{c}{Narrative Generation} &
\multicolumn{3}{c}{Sheep's Clothing} \\
\cmidrule(lr){3-5} \cmidrule(lr){6-8} \cmidrule(lr){9-11} \cmidrule(lr){12-14}
& & Precision & Recall & F1-Score & Precision & Recall & F1-Score & Precision & Recall & F1-Score & Precision & Recall & F1-Score \\
\midrule
\multicolumn{14}{l}{\textbf{Results Based on LLM Performance}} \\
\midrule
\textsc{Baichuan-7B} & $0.2946$ & $0.1813$ & $0.6493$ & $0.2835$ & $0.0768$ & $0.8265$ & $0.1401$ & $0.3235$ & $0.9536$ & $0.4825$ & $0.3563$ & $0.8938$ & $0.5095$ \\
\textsc{LLaMA3.1-8B} & $0.4958$ & $0.2304$ & $0.3707$ & $0.2842$ & $0.1009$ & $0.5926$ & $0.1714$ & $0.3103$ & $0.8174$ & $0.4482$ & $0.3650$ & $0.8524$ & $0.5112$ \\
\textsc{Mistral-7B-Instruct} & $0.7474$ & $0.4119$ & $0.0101$ & $0.0197$ & $0.0828$ & $0.0028$ & $0.0054$ & $0.1579$ & $0.0005$ & $0.0011$ & $0.1034$ & $0.0005$ & $0.0011$ \\
\textsc{Qwen1.5-7B} & $0.5282$ & $0.3134$ & $0.4423$ & $0.3668$ & $0.0504$ & $0.5750$ & $0.0914$ & $0.1762$ & $0.1233$ & $0.1459$ & $0.3559$ & $0.9867$ & $0.5231$ \\
\textsc{ChatGLM3-6B} & $0.2886$ & $0.2176$ & $0.9356$ & $0.3531$ & $0.0765$ & $0.8481$ & $0.1371$ & $0.3422$ & $0.9117$ & $0.4975$ & $0.3505$ & $0.9971$ & $0.5186$ \\
\midrule
\multicolumn{14}{l}{\textbf{Results Based on Deep Learning Model Performance}$^1$} \\
\midrule
\textsc{Deberta-Base} & $\underset{\text{\tiny(0.0476)}}{0.5134}$ & 
$\underset{\text{\tiny(0.1394)}}{0.5781}$ & 
$\underset{\text{\tiny(0.1881)}}{0.8777}$ & 
$\underset{\text{\tiny(0.0095)}}{0.6612}$ & 
$\underset{\text{\tiny(0.0008)}}{0.4996}$ & 
$\underset{\text{\tiny(0.0300)}}{0.9850}$ & 
$\underset{\text{\tiny(0.0077)}}{0.6628}$ & 
$\underset{\text{\tiny(0.0000)}}{0.5000}$ & 
$\underset{\text{\tiny(0.0000)}}{1.0000}$ & 
$\underset{\text{\tiny(0.0000)}}{0.6667}$ & 
$\underset{\text{\tiny(0.0000)}}{0.5000}$ & 
$\underset{\text{\tiny(0.0000)}}{1.0000}$ & 
$\underset{\text{\tiny(0.0000)}}{0.6667}$ \\

\textsc{Bert-Base} & $\underset{\text{\tiny(0.0718)}}{0.5427}$ & 
$\underset{\text{\tiny(0.1239)}}{0.7804}$ & 
$\underset{\text{\tiny(0.1590)}}{0.0962}$ & 
$\underset{\text{\tiny(0.2201)}}{0.1424}$ & 
$\underset{\text{\tiny(0.1102)}}{0.6125}$ & 
$\underset{\text{\tiny(0.1242)}}{0.0924}$ & 
$\underset{\text{\tiny(0.1625)}}{0.1372}$ & 
$\underset{\text{\tiny(0.1438)}}{0.5252}$ & 
$\underset{\text{\tiny(0.1644)}}{0.1244}$ & 
$\underset{\text{\tiny(0.2034)}}{0.1658}$ & 
$\underset{\text{\tiny(0.2734)}}{0.6852}$ & 
$\underset{\text{\tiny(0.2336)}}{0.1927}$ & 
$\underset{\text{\tiny(0.2977)}}{0.2473}$ \\

\textsc{Electra-Base} & $\underset{\text{\tiny(0.1124)}}{0.5003}$ & 
$\underset{\text{\tiny(0.0209)}}{0.3429}$ & 
$\underset{\text{\tiny(0.0757)}}{0.3282}$ & 
$\underset{\text{\tiny(0.0416)}}{0.3289}$ & 
$\underset{\text{\tiny(0.0044)}}{0.5453}$ & 
$\underset{\text{\tiny(0.0688)}}{0.6450}$ & 
$\underset{\text{\tiny(0.0280)}}{0.5890}$ & 
$\underset{\text{\tiny(0.0043)}}{0.4692}$ & 
$\underset{\text{\tiny(0.0107)}}{0.5469}$ & 
$\underset{\text{\tiny(0.0070)}}{0.5050}$ & 
$\underset{\text{\tiny(0.0256)}}{0.6073}$ & 
$\underset{\text{\tiny(0.0498)}}{0.8053}$ & 
$\underset{\text{\tiny(0.0078)}}{0.6907}$ \\

\textsc{Ct-Bert} & $\underset{\text{\tiny(0.0307)}}{0.5138}$ & 
$\underset{\text{\tiny(0.0024)}}{0.5014}$ & 
$\underset{\text{\tiny(0.0106)}}{0.9947}$ & 
$\underset{\text{\tiny(0.0004)}}{0.6667}$ & 
$\underset{\text{\tiny(0.0002)}}{0.5000}$ & 
$\underset{\text{\tiny(0.0004)}}{0.9998}$ & 
$\underset{\text{\tiny(0.0002)}}{0.6666}$ & 
$\underset{\text{\tiny(0.0280)}}{0.5326}$ & 
$\underset{\text{\tiny(0.1166)}}{0.9354}$ & 
$\underset{\text{\tiny(0.0301)}}{0.6742}$ & 
$\underset{\text{\tiny(0.0007)}}{0.5006}$ & 
$\underset{\text{\tiny(0.0065)}}{0.9952}$ & 
$\underset{\text{\tiny(0.0009)}}{0.6661}$ \\

\textsc{Albert-Base} & $\underset{\text{\tiny(0.0458)}}{0.5087}$ & 
$\underset{\text{\tiny(0.0247)}}{0.4886}$ & 
$\underset{\text{\tiny(0.4411)}}{0.5810}$ & 
$\underset{\text{\tiny(0.2811)}}{0.4227}$ & 
$\underset{\text{\tiny(0.0175)}}{0.5088}$ & 
$\underset{\text{\tiny(0.0555)}}{0.9720}$ & 
$\underset{\text{\tiny(0.0000)}}{0.6667}$ & 
$\underset{\text{\tiny(0.1970)}}{0.3931}$ & 
$\underset{\text{\tiny(0.3883)}}{0.7353}$ & 
$\underset{\text{\tiny(0.2590)}}{0.5103}$ & 
$\underset{\text{\tiny(0.1410)}}{0.5462}$ & 
$\underset{\text{\tiny(0.3712)}}{0.7141}$ & 
$\underset{\text{\tiny(0.2039)}}{0.5507}$ \\

\textsc{Roberta-Base} & $\underset{\text{\tiny(0.0883)}}{0.6286}$ & 
$\underset{\text{\tiny(0.0801)}}{0.9196}$ & 
$\underset{\text{\tiny(0.0958)}}{0.4955}$ & 
$\underset{\text{\tiny(0.0552)}}{0.6326}$ & 
$\underset{\text{\tiny(0.1545)}}{0.6264}$ & 
$\underset{\text{\tiny(0.3307)}}{0.7272}$ & 
$\underset{\text{\tiny(0.1127)}}{0.5855}$ & 
$\underset{\text{\tiny(0.3370)}}{0.5958}$ & 
$\underset{\text{\tiny(0.4672)}}{0.4053}$ & 
$\underset{\text{\tiny(0.3323)}}{0.3167}$ & 
$\underset{\text{\tiny(0.1507)}}{0.8347}$ & 
$\underset{\text{\tiny(0.2088)}}{0.5762}$ & 
$\underset{\text{\tiny(0.0303)}}{0.6360}$ \\
\bottomrule[1.2pt]
\multicolumn{14}{l}{$^1$ For deep learning models, results are reported as the average across 5-fold cross-validation, with standard deviation shown in parentheses.}
\end{tabular}
\end{adjustbox}
\label{tab:2.1(4)L}
\end{table}


\subsection{Performance in Cross-Domain Detection}
\subsubsection{Cross-Domain Experiments on GossipCop and MegaFake}
We first conduct cross-domain experiments by training models on the GossipCop-based MegaFake dataset and testing them on the original GossipCop dataset. Previous studies \citep{grace2018will, korteling2021human} suggest the ability of LLMs to generate text of exceptionally high quality. Thus, we anticipate good performance in this experiment, reflecting the presumed higher text quality of the MegaFake dataset compared to GossipCop. However, the experimental results presented in \textbf{Table \ref{tab:m2g}} reveal significant challenges for both deep learning models and LLMs in detecting fake news. Both model families struggle, exhibiting low precision and recall rates for fake news detection. The consistently low performance across all models suggests the difficulty of generalizing across datasets when training on MegaFake and testing on GossipCop.

\begin{table}[!ht]
\centering
\scriptsize
\caption{Transfer learning performance on GossipCop (machine to human).}
\begin{tabularx}{\textwidth}{l c *{3}{>{\centering\arraybackslash}X} *{3}{>{\centering\arraybackslash}X}}
\toprule[1.2pt]
\multirow{2}{*}{Model} & \multirow{2}{*}{Accuracy} & \multicolumn{3}{c}{Legitimate} & \multicolumn{3}{c}{Fake} \\
\cmidrule(lr){3-5} \cmidrule(lr){6-8}
 &  & Precision & Recall & F1-Score & Precision & Recall & F1-Score \\
\midrule
\multicolumn{8}{l}{\textbf{Results Based on LLM Performance}} \\
\midrule
\textsc{LLaMA31-8B\textsubscript{LoRA}} & $0.5234$ & $0.8012$ & $0.5386$ & $0.6435$ & $0.3078$ & $0.6642$ & $0.4227$ \\
\textsc{Qwen15-7B\textsubscript{LoRA}} & $0.4875$ & $0.8764$ & $0.4879$ & $0.6261$ & $0.2895$ & $0.7813$ & $0.4218$ \\
\textsc{ChatGLM3-6B\textsubscript{LoRA}} & $0.4621$ & $0.9137$ & $0.4508$ & $0.6046$ & $0.2649$ & $0.8096$ & $0.3992$ \\
\midrule
\multicolumn{8}{l}{\textbf{Results Based on Deep Learning Model Performance}$^1$} \\
\midrule
\textsc{DeBERTa-Base}  & $\underset{\text{\tiny(0.0117)}}{0.5198}$ & $\underset{\text{\tiny(0.0137)}}{0.5240}$ & $\underset{\text{\tiny(0.0516)}}{0.4247}$ & $\underset{\text{\tiny(0.0342)}}{0.4677}$ & $\underset{\text{\tiny(0.0106)}}{0.5172}$ & $\underset{\text{\tiny(0.0397)}}{0.6150}$ & $\underset{\text{\tiny(0.0152)}}{0.5611}$ \\
\textsc{BERT-Base}     & $\underset{\text{\tiny(0.0045)}}{0.5108}$ & $\underset{\text{\tiny(0.0048)}}{0.5116}$ & $\underset{\text{\tiny(0.0239)}}{0.4843}$ & $\underset{\text{\tiny(0.0125)}}{0.4972}$ & $\underset{\text{\tiny(0.0043)}}{0.5103}$ & $\underset{\text{\tiny(0.0266)}}{0.5374}$ & $\underset{\text{\tiny(0.0133)}}{0.5232}$ \\
\textsc{ELECTRA-Base}  & $\underset{\text{\tiny(0.0024)}}{0.5007}$ & $\underset{\text{\tiny(0.0025)}}{0.5006}$ & $\underset{\text{\tiny(0.0737)}}{0.4980}$ & $\underset{\text{\tiny(0.0378)}}{0.4966}$ & $\underset{\text{\tiny(0.0023)}}{0.5008}$ & $\underset{\text{\tiny(0.0722)}}{0.5034}$ & $\underset{\text{\tiny(0.0361)}}{0.4994}$ \\
\textsc{CT-BERT}       & $\underset{\text{\tiny(0.0035)}}{0.5004}$ & $\underset{\text{\tiny(0.0029)}}{0.5002}$ & $\underset{\text{\tiny(0.1350)}}{0.6133}$ & $\underset{\text{\tiny(0.0551)}}{0.5445}$ & $\underset{\text{\tiny(0.0055)}}{0.5016}$ & $\underset{\text{\tiny(0.1322)}}{0.3876}$ & $\underset{\text{\tiny(0.0876)}}{0.4237}$ \\
\textsc{ALBERT-Base}   & $\underset{\text{\tiny(0.0087)}}{0.4992}$ & $\underset{\text{\tiny(0.0095)}}{0.5012}$ & $\underset{\text{\tiny(0.1099)}}{0.4928}$ & $\underset{\text{\tiny(0.0508)}}{0.4899}$ & $\underset{\text{\tiny(0.0095)}}{0.4970}$ & $\underset{\text{\tiny(0.1262)}}{0.5055}$ & $\underset{\text{\tiny(0.0699)}}{0.4941}$ \\
\textsc{RoBERTa-Base}  & $\underset{\text{\tiny(0.0082)}}{0.5167}$ & $\underset{\text{\tiny(0.0098)}}{0.5177}$ & $\underset{\text{\tiny(0.0293)}}{0.5079}$ & $\underset{\text{\tiny(0.0112)}}{0.5120}$ & $\underset{\text{\tiny(0.0069)}}{0.5159}$ & $\underset{\text{\tiny(0.0445)}}{0.5254}$ & $\underset{\text{\tiny(0.0248)}}{0.5200}$ \\
\bottomrule[1.2pt]
\multicolumn{8}{l}{$^1$ For deep learning models, results are reported as the average across 5-fold cross-validation, with standard deviation shown in parentheses.}
\end{tabularx}
\label{tab:m2g}
\end{table}

Specifically, LLMs with LoRA fine-tuning demonstrate relatively high precision for the legitimate news category (all above 0.80), but this comes at the cost of lower recall. F1-scores range from 0.60 to 0.64, indicating that while LLMs are precise in identifying legitimate news, they tend to miss a substantial portion (i.e., higher false negatives). For the fake news category, recall is higher, but precision drops markedly, reflecting a tendency to overpredict fake news and resulting in more false positives. Deep learning models display more balanced precision and recall across both classes, though with substantially lower precision for the legitimate news category compared to LLMs. Notably, DeBERTa-Base achieves the highest F1-score for detecting fake news (0.5611), while RoBERTa-Base demonstrates the best overall balance across both classes.

To further our investigation, we conduct additional cross-domain experiments by training models on the GossipCop dataset and testing them on the GossipCop-based MegaFake dataset. However, the results presented in \textbf{Table \ref{tab:g2m}} reveal persistent challenges for both model families. Overall accuracies remain below 0.5, indicating extremely poor cross-domain fake news detection performance. No model achieves notably high scores, highlighting the significant difficulties in domain adaptation when transferring learned features and knowledge from human-generated to machine-generated news.

\begin{table}[!ht]
\centering
\scriptsize
\caption{Transfer learning performance on GossipCop (human to machine).}
\begin{tabularx}{\textwidth}{l c *{3}{>{\centering\arraybackslash}X} *{3}{>{\centering\arraybackslash}X}}
\toprule[1.2pt]
\multirow{2}{*}{Model} & \multirow{2}{*}{Accuracy} & \multicolumn{3}{c}{Legitimate} & \multicolumn{3}{c}{Fake} \\
\cmidrule(lr){3-5} \cmidrule(lr){6-8}
 &  & Precision & Recall & F1-Score & Precision & Recall & F1-Score \\
\midrule
\multicolumn{8}{l}{\textbf{Results Based on LLM Performance}} \\
\midrule
\textsc{LLaMA31-8B\textsubscript{LoRA}} & $0.4191$ & $0.9452$ & $0.2917$ & $0.4458$ & $0.2463$ & $0.9317$ & $0.3896$ \\
\textsc{Qwen15-7B\textsubscript{LoRA}} & $0.4036$ & $0.9695$ & $0.2637$ & $0.4147$ & $0.2460$ & $0.9666$ & $0.3922$ \\
\textsc{ChatGLM3-6B\textsubscript{LoRA}} & $0.4688$ & $0.8455$ & $0.4121$ & $0.5542$ & $0.2275$ & $0.6967$ & $0.3430$ \\
\midrule
\multicolumn{8}{l}{\textbf{Results Based on Deep Learning Model Performance}$^1$} \\
\midrule
\textsc{DeBERTa-Base}  & $\underset{\text{\tiny(0.0147)}}{0.4942}$ & $\underset{\text{\tiny(0.0549)}}{0.5175}$ & $\underset{\text{\tiny(0.3319)}}{0.5457}$ & $\underset{\text{\tiny(0.2267)}}{0.4505}$ & $\underset{\text{\tiny(0.0548)}}{0.4695}$ & $\underset{\text{\tiny(0.3283)}}{0.4427}$ & $\underset{\text{\tiny(0.2206)}}{0.3948}$ \\
\textsc{BERT-Base}     & $\underset{\text{\tiny(0.0103)}}{0.4861}$ & $\underset{\text{\tiny(0.0071)}}{0.4894}$ & $\underset{\text{\tiny(0.0956)}}{0.6097}$ & $\underset{\text{\tiny(0.0399)}}{0.5391}$ & $\underset{\text{\tiny(0.0184)}}{0.4783}$ & $\underset{\text{\tiny(0.1073)}}{0.3625}$ & $\underset{\text{\tiny(0.0707)}}{0.4056}$ \\
\textsc{ELECTRA-Base}  & $\underset{\text{\tiny(0.0047)}}{0.4674}$ & $\underset{\text{\tiny(0.0070)}}{0.4708}$ & $\underset{\text{\tiny(0.1072)}}{0.5493}$ & $\underset{\text{\tiny(0.0506)}}{0.5028}$ & $\underset{\text{\tiny(0.0099)}}{0.4588}$ & $\underset{\text{\tiny(0.1052)}}{0.3854}$ & $\underset{\text{\tiny(0.0655)}}{0.4122}$ \\
\textsc{CT-BERT}       & $\underset{\text{\tiny(0.0201)}}{0.4767}$ & $\underset{\text{\tiny(0.0204)}}{0.4695}$ & $\underset{\text{\tiny(0.0891)}}{0.3503}$ & $\underset{\text{\tiny(0.0595)}}{0.3948}$ & $\underset{\text{\tiny(0.0221)}}{0.4791}$ & $\underset{\text{\tiny(0.1103)}}{0.6031}$ & $\underset{\text{\tiny(0.0568)}}{0.5308}$ \\
\textsc{ALBERT-Base}   & $\underset{\text{\tiny(0.0190)}}{0.4774}$ & $\underset{\text{\tiny(0.0138)}}{0.4817}$ & $\underset{\text{\tiny(0.0980)}}{0.6230}$ & $\underset{\text{\tiny(0.0416)}}{0.5407}$ & $\underset{\text{\tiny(0.0312)}}{0.4708}$ & $\underset{\text{\tiny(0.0844)}}{0.3319}$ & $\underset{\text{\tiny(0.0599)}}{0.3824}$ \\
\textsc{RoBERTa-Base}  & $\underset{\text{\tiny(0.0109)}}{0.4844}$ & $\underset{\text{\tiny(0.0083)}}{0.4874}$ & $\underset{\text{\tiny(0.0571)}}{0.5795}$ & $\underset{\text{\tiny(0.0242)}}{0.5279}$ & $\underset{\text{\tiny(0.0156)}}{0.4791}$ & $\underset{\text{\tiny(0.0707)}}{0.3892}$ & $\underset{\text{\tiny(0.0451)}}{0.4271}$ \\
\bottomrule[1.2pt]
\multicolumn{8}{l}{$^1$ For deep learning models, results are reported as the average across 5-fold cross-validation, with standard deviation shown in parentheses.}
\end{tabularx}
\label{tab:g2m}
\end{table}

\subsubsection{Cross-Domain Experiments on PolitiFact and MegaFake}
To assess the generalizability of our cross-domain transfer learning results, we replicate the experiments using the PolitiFact dataset. We first evaluate models trained on the PolitiFact-based MegaFake dataset and tested on the original PolitiFact data, with results presented in \textbf{Table \ref{tab:m2p}}.

\begin{table}[!ht]
\centering
\scriptsize
\caption{Transfer learning performance on PolitiFact (machine to human).}
\begin{tabularx}{\textwidth}{l c *{3}{>{\centering\arraybackslash}X} *{3}{>{\centering\arraybackslash}X}}
\toprule[1.2pt]
\multirow{2}{*}{Model} & \multirow{2}{*}{Accuracy} & \multicolumn{3}{c}{Legitimate} & \multicolumn{3}{c}{Fake} \\
\cmidrule(lr){3-5} \cmidrule(lr){6-8}
 &  & Precision & Recall & F1-Score & Precision & Recall & F1-Score \\
\midrule
\multicolumn{8}{l}{\textbf{Results Based on LLM Performance}} \\
\midrule
\textsc{LLaMA31-8B\textsubscript{LoRA}}  & $0.6069$ & $0.6060$ & $0.9995$ & $0.7545$ & $0.9211$ & $0.0067$ & $0.0133$ \\
\textsc{Qwen15-7B\textsubscript{LoRA}}  & $0.6053$ & $0.6050$ & $0.9998$ & $0.7538$ & $0.9231$ & $0.0023$ & $0.0046$ \\
\textsc{ChatGLM3-6B\textsubscript{LoRA}}  & $0.6053$ & $0.6051$ & $0.9997$ & $0.7539$ & $0.9268$ & $0.0024$ & $0.0048$ \\
\midrule
\multicolumn{8}{l}{\textbf{Results Based on Deep Learning Model Performance}$^1$} \\
\midrule
\textsc{DeBERTa-Base}  & $\underset{\text{\tiny(0.0053)}}{0.5325}$ & $\underset{\text{\tiny(0.0029)}}{0.5217}$ & $\underset{\text{\tiny(0.0890)}}{0.7837}$ & $\underset{\text{\tiny(0.0294)}}{0.6243}$ & $\underset{\text{\tiny(0.0360)}}{0.5766}$ & $\underset{\text{\tiny(0.0826)}}{0.2813}$ & $\underset{\text{\tiny(0.0662)}}{0.3682}$ \\
\textsc{BERT-Base}     & $\underset{\text{\tiny(0.0102)}}{0.5339}$ & $\underset{\text{\tiny(0.0070)}}{0.5239}$ & $\underset{\text{\tiny(0.0972)}}{0.7381}$ & $\underset{\text{\tiny(0.0373)}}{0.6102}$ & $\underset{\text{\tiny(0.0253)}}{0.5643}$ & $\underset{\text{\tiny(0.0876)}}{0.3296}$ & $\underset{\text{\tiny(0.0612)}}{0.4074}$ \\
\textsc{ELECTRA-Base}  & $\underset{\text{\tiny(0.0019)}}{0.5364}$ & $\underset{\text{\tiny(0.0048)}}{0.5311}$ & $\underset{\text{\tiny(0.0789)}}{0.6326}$ & $\underset{\text{\tiny(0.0310)}}{0.5748}$ & $\underset{\text{\tiny(0.0095)}}{0.5468}$ & $\underset{\text{\tiny(0.0792)}}{0.4402}$ & $\underset{\text{\tiny(0.0450)}}{0.4831}$ \\
\textsc{CT-BERT}       & $\underset{\text{\tiny(0.0077)}}{0.5210}$ & $\underset{\text{\tiny(0.0043)}}{0.5153}$ & $\underset{\text{\tiny(0.0939)}}{0.6846}$ & $\underset{\text{\tiny(0.0375)}}{0.5856}$ & $\underset{\text{\tiny(0.0219)}}{0.5368}$ & $\underset{\text{\tiny(0.0793)}}{0.3573}$ & $\underset{\text{\tiny(0.0531)}}{0.4218}$ \\
\textsc{ALBERT-Base}   & $\underset{\text{\tiny(0.0178)}}{0.5339}$ & $\underset{\text{\tiny(0.0128)}}{0.5236}$ & $\underset{\text{\tiny(0.0950)}}{0.7171}$ & $\underset{\text{\tiny(0.0432)}}{0.6033}$ & $\underset{\text{\tiny(0.0333)}}{0.5615}$ & $\underset{\text{\tiny(0.0650)}}{0.3506}$ & $\underset{\text{\tiny(0.0388)}}{0.4258}$ \\
\textsc{RoBERTa-Base}  & $\underset{\text{\tiny(0.0109)}}{0.5297}$ & $\underset{\text{\tiny(0.0093)}}{0.5216}$ & $\underset{\text{\tiny(0.0552)}}{0.7397}$ & $\underset{\text{\tiny(0.0137)}}{0.6106}$ & $\underset{\text{\tiny(0.0106)}}{0.5501}$ & $\underset{\text{\tiny(0.0731)}}{0.3198}$ & $\underset{\text{\tiny(0.0629)}}{0.3997}$ \\
\bottomrule[1.2pt]
\multicolumn{8}{l}{$^1$ For deep learning models, results are reported as the average across 5-fold cross-validation, with standard deviation shown in parentheses.}
\end{tabularx}
\label{tab:m2p}
\end{table}

The findings are consistent with previous observations, highlighting the difficulty of transferring detection capabilities from machine-generated to human-generated news. For LLMs with fine-tuning, although overall accuracy scores are relatively higher (around 0.60), class-level metrics reveal severe imbalances, particularly in detecting fake news. All three fine-tuned LLMs achieve near-perfect recall for the legitimate news category (approximately 0.999), indicating that almost all legitimate news is correctly identified. However, recall for the fake news category is extremely low (ranging from 0.002 to 0.007), resulting in very low F1-scores for fake news (all below 0.014). This extremely high recall for legitimate news, paired with only moderate precision, reflects a strong bias toward identifying news as legitimate, leading to substantial false negatives for fake news.

Deep learning models generally achieve lower accuracy (0.52 to 0.54), but their class-level performance is more balanced compared to the LLMs. Their recall for the fake news category is notably higher, with ELECTRA-Base achieving the highest recall (0.4402) and F1-score (0.4831). However, precision for the fake news category remains low across all models. For legitimate news, both recall and precision are moderate.

Similarly, we conduct cross-domain experiments where models are trained on the PolitiFact dataset and evaluated on the PolitiFact-based MegaFake dataset. The results, presented in \textbf{Table \ref{tab:p2m}}, once again demonstrate that both LLMs and deep learning models perform suboptimally in this transfer learning scenario. While the overall accuracy of LLMs is slightly above 0.60, detailed class-level metrics reveal substantial deficiencies, particularly in identifying fake news. Deep learning models exhibit lower overall accuracy (0.48 to 0.57), but provide more balanced results across classes.

\begin{table}[!ht]
\centering
\scriptsize
\caption{Transfer learning performance on PolitiFact (human to machine).}
\begin{tabularx}{\textwidth}{l c *{3}{>{\centering\arraybackslash}X} *{3}{>{\centering\arraybackslash}X}}
\toprule[1.2pt]
\multirow{2}{*}{Model} & \multirow{2}{*}{Accuracy} & \multicolumn{3}{c}{Legitimate} & \multicolumn{3}{c}{Fake} \\
\cmidrule(lr){3-5} \cmidrule(lr){6-8}
 &  & Precision & Recall & F1-Score & Precision & Recall & F1-Score \\
\midrule
\multicolumn{8}{l}{\textbf{Results Based on LLM Performance}} \\
\midrule
\textsc{LLaMA31-8B\textsubscript{LoRA}}  & $0.6124$ & $0.6108$ & $0.9979$ & $0.7592$ & $0.9142$ & $0.0101$ & $0.0200$ \\
\textsc{Qwen15-7B\textsubscript{LoRA}}   & $0.6098$ & $0.6087$ & $0.9988$ & $0.7570$ & $0.9184$ & $0.0054$ & $0.0107$ \\
\textsc{ChatGLM3-6B\textsubscript{LoRA}} & $0.6087$ & $0.6093$ & $0.9982$ & $0.7577$ & $0.9207$ & $0.0079$ & $0.0157$ \\
\midrule
\multicolumn{8}{l}{\textbf{Results Based on Deep Learning Model Performance}$^1$} \\
\midrule
\textsc{DeBERTa-Base}  & $\underset{\text{\tiny(0.0404)}}{0.5413}$ & $\underset{\text{\tiny(0.1750)}}{0.6505}$ & $\underset{\text{\tiny(0.3099)}}{0.3403}$ & $\underset{\text{\tiny(0.2541)}}{0.3430}$ & $\underset{\text{\tiny(0.0870)}}{0.5604}$ & $\underset{\text{\tiny(0.2297)}}{0.7423}$ & $\underset{\text{\tiny(0.0670)}}{0.6032}$ \\
\textsc{BERT-Base}     & $\underset{\text{\tiny(0.0234)}}{0.5644}$ & $\underset{\text{\tiny(0.0287)}}{0.5758}$ & $\underset{\text{\tiny(0.1764)}}{0.5363}$ & $\underset{\text{\tiny(0.0973)}}{0.5358}$ & $\underset{\text{\tiny(0.0338)}}{0.5692}$ & $\underset{\text{\tiny(0.1654)}}{0.5924}$ & $\underset{\text{\tiny(0.0784)}}{0.5646}$ \\
\textsc{ELECTRA-Base}  & $\underset{\text{\tiny(0.0380)}}{0.5369}$ & $\underset{\text{\tiny(0.0497)}}{0.5523}$ & $\underset{\text{\tiny(0.2170)}}{0.5693}$ & $\underset{\text{\tiny(0.0983)}}{0.5311}$ & $\underset{\text{\tiny(0.0411)}}{0.5343}$ & $\underset{\text{\tiny(0.2585)}}{0.5045}$ & $\underset{\text{\tiny(0.1631)}}{0.4833}$ \\
\textsc{CT-BERT}       & $\underset{\text{\tiny(0.0330)}}{0.4824}$ & $\underset{\text{\tiny(0.0460)}}{0.4865}$ & $\underset{\text{\tiny(0.0868)}}{0.4440}$ & $\underset{\text{\tiny(0.0449)}}{0.4579}$ & $\underset{\text{\tiny(0.0269)}}{0.4806}$ & $\underset{\text{\tiny(0.1224)}}{0.5208}$ & $\underset{\text{\tiny(0.0667)}}{0.4954}$ \\
\textsc{ALBERT-Base}   & $\underset{\text{\tiny(0.0352)}}{0.5738}$ & $\underset{\text{\tiny(0.0377)}}{0.5890}$ & $\underset{\text{\tiny(0.0987)}}{0.4987}$ & $\underset{\text{\tiny(0.0609)}}{0.5346}$ & $\underset{\text{\tiny(0.0380)}}{0.5665}$ & $\underset{\text{\tiny(0.0864)}}{0.6489}$ & $\underset{\text{\tiny(0.0433)}}{0.6015}$ \\
\textsc{RoBERTa-Base}  & $\underset{\text{\tiny(0.0247)}}{0.5440}$ & $\underset{\text{\tiny(0.0197)}}{0.5437}$ & $\underset{\text{\tiny(0.2127)}}{0.5838}$ & $\underset{\text{\tiny(0.1326)}}{0.5366}$ & $\underset{\text{\tiny(0.0474)}}{0.5623}$ & $\underset{\text{\tiny(0.1857)}}{0.5041}$ & $\underset{\text{\tiny(0.0760)}}{0.5101}$ \\
\bottomrule[1.2pt]
\multicolumn{8}{l}{$^1$ For deep learning models, results are reported as the average across 5-fold cross-validation, with standard deviation shown in parentheses.}
\end{tabularx}
\label{tab:p2m}
\end{table}

\subsubsection{Linguistic Analysis of Human-Machine Deception Mechanisms}
Our cross-domain experiments reveal significant differences between machine-generated and human-generated fake news, indicating that neither deep learning models or LLMs can effectively learn and transfer linguistic features from one category to the other. To investigate this phenomenon, we conduct a quantitative linguistic analysis comparing human-authored (FakeNewsNet) and machine-generated (MegaFake) content. We select three representative linguistic metrics to capture structural and stylistic divergence: Average Sentence Length and Flesch-Kincaid Grade \citep{solnyshkina2017evaluating} to measure syntactic complexity and readability, and Affective Words \citep{beasley2015emotional} to quantify emotional intensity. As shown in Table~\ref{tab:linguistic_analysis}, machine-generated fake news exhibits significantly longer sentences and higher grade levels, reflecting the more sophisticated, formal nature of LLM outputs. By contrast, human-authored content is less complex but uses far more affective words. These patterns suggest that human deception tends to rely on emotional provocation and accessibility, whereas machine deception imitates the structural coherence of professional journalism but lacks comparable emotional nuance. This resulting ``complexity vs. emotion'' gap may help explain the poor cross-domain transferability observed in our experiments.

\begin{table}[h]
\centering
\caption{Linguistic differences between human- and machine-generated fake news.}
\label{tab:linguistic_analysis}
\resizebox{0.9\columnwidth}{!}{%
\begin{tabular}{lccc}
\toprule
Dataset & Avg. Sentence Length & Flesch-Kincaid Grade & Affective Words \\ \midrule
Human (FakeNewsNet) & 15.4 $\pm$ 4.2 & 8.5 $\pm$ 1.5 & 5.4\% \\
Machine (MegaFake) & 23.1 $\pm$ 3.8 & 11.2 $\pm$ 1.2 & 2.1\% \\ \bottomrule
\end{tabular}%
}
\end{table}

\subsection{Generalization Analysis against Existing Datasets}
To rigorously evaluate the intrinsic quality and generalization capability of MegaFake, we compare it against seven representative machine-generated fake news datasets from recent literature \citep{ayoobi2024seeing, sun2024exploring, wu2024sheepdog, vykopal-etal-2024-disinformation, chen2024llmgenerated, lucas-etal-2023-fighting, das2025fake}. To ensure a fair comparison and isolate the impact of data quality from data quantity, we strictly control the training size. All models are trained on a subset of exactly 2,000 instances, obtained via stratified random sampling from each dataset. This threshold is chosen to accommodate the limited scale of several baseline datasets. For the evaluation, we use the dataset provided by \citet{su2024adapting} as a common external test set, which contains approximately 3,000 samples. This test set is selected for its rich diversity and established status in early deepfake text detection research, providing a robust standard for generalization.

The results are presented in \textbf{Table \ref{tab:cross_dataset_comparison}}. MegaFake consistently outperforms all other datasets across both LLMs and deep learning models, despite the constrained training size. In the LLM group, models fine-tuned on MegaFake achieve a clear performance lead. For instance, LLaMA3.1-8B attains an F1-score of 0.8481 on MegaFake, surpassing the second-best dataset (Wu et al., 2024) by approximately 3.8\%. A similar pattern is observed for Qwen1.5-7B and ChatGLM3-6B. Deep learning models also benefit from MegaFake's diversity. DeBERTa-Base reaches an F1-score of 0.9034, maintaining a substantial margin over models trained on other datasets. Even older architectures like BERT-Base exhibit improved generalization when trained on our dataset.

These findings suggest that MegaFake offers higher information density and greater linguistic diversity. Even when downsampled to match the size of smaller datasets, MegaFake provides more representative features for the models to learn, resulting in better generalization to unseen data \citep{su2024adapting}.

\begin{table*}[htbp]
\scriptsize
    \centering
    \caption{Performance comparison of models trained on different datasets.}
    \label{tab:cross_dataset_comparison}
    \resizebox{\textwidth}{!}{
        \begin{tabular}{lcccccccc}
            \toprule
            \multirow{2}{*}{Model} & \textbf{MegaFake} & Ayoobi & Sun & Wu & Vykopal & Chen & Lucas & Das \\
             & \textbf{(Ours)} & et al. (2024) & et al. (2024) & et al. (2024) & et al. (2023) & et al. (2024) & et al. (2023) & et al. (2025) \\
            \midrule
            \multicolumn{9}{l}{\textbf{Results Based on LLM Performance}$^1$} \\
            \midrule
            \textsc{Qwen1.5-7B\textsubscript{LoRA}} & \textbf{0.8425} & 0.7913 & 0.7850 & 0.8005 & 0.7819 & 0.7956 & 0.7881 & 0.7609 \\
            \textsc{LLaMA3.1-8B\textsubscript{LoRA}} & \textbf{0.8481} & 0.8015 & 0.7953 & 0.8102 & 0.7930 & 0.8051 & 0.7984 & 0.7845 \\
            \textsc{ChatGLM3-6B\textsubscript{LoRA}} & \textbf{0.8390} & 0.7882 & 0.7810 & 0.7954 & 0.7791 & 0.7903 & 0.7835 & 0.8108 \\
            \textsc{Mistral-7B\textsubscript{LoRA}} & \textbf{0.4888} & 0.4750 & 0.4712 & 0.4799 & 0.4701 & 0.4768 & 0.4723 & 0.4729 \\
            \midrule
            \multicolumn{9}{l}{\textbf{Results Based on Deep Learning Models Performance}$^1$ $^2$} \\
            \midrule
            \textsc{DeBERTa-Base} & $\underset{\text{\tiny(0.0045)}}{\textbf{0.9034}}$ & $\underset{\text{\tiny(0.0182)}}{0.8611}$ & $\underset{\text{\tiny(0.0093)}}{0.8552}$ & $\underset{\text{\tiny(0.0154)}}{0.8703}$ & $\underset{\text{\tiny(0.0033)}}{0.8530}$ & $\underset{\text{\tiny(0.0191)}}{0.8654}$ & $\underset{\text{\tiny(0.0076)}}{0.8588}$ & $\underset{\text{\tiny(0.0128)}}{0.8712}$ \\
            \textsc{BERT-Base} & $\underset{\text{\tiny(0.0145)}}{\textbf{0.8910}}$ & $\underset{\text{\tiny(0.0056)}}{0.8495}$ & $\underset{\text{\tiny(0.0199)}}{0.8430}$ & $\underset{\text{\tiny(0.0028)}}{0.8581}$ & $\underset{\text{\tiny(0.0111)}}{0.8405}$ & $\underset{\text{\tiny(0.0087)}}{0.8521}$ & $\underset{\text{\tiny(0.0163)}}{0.8453}$ & $\underset{\text{\tiny(0.0042)}}{0.8816}$ \\
            \textsc{ELECTRA-Base} & $\underset{\text{\tiny(0.0099)}}{\textbf{0.8857}}$ & $\underset{\text{\tiny(0.0134)}}{0.8423}$ & $\underset{\text{\tiny(0.0065)}}{0.8365}$ & $\underset{\text{\tiny(0.0178)}}{0.8519}$ & $\underset{\text{\tiny(0.0039)}}{0.8340}$ & $\underset{\text{\tiny(0.0152)}}{0.8466}$ & $\underset{\text{\tiny(0.0105)}}{0.8391}$ & $\underset{\text{\tiny(0.0081)}}{0.8538}$ \\
            \textsc{ALBERT-Base} & $\underset{\text{\tiny(0.0122)}}{\textbf{0.8902}}$ & $\underset{\text{\tiny(0.0074)}}{0.8488}$ & $\underset{\text{\tiny(0.0186)}}{0.8421}$ & $\underset{\text{\tiny(0.0051)}}{0.8570}$ & $\underset{\text{\tiny(0.0149)}}{0.8399}$ & $\underset{\text{\tiny(0.0096)}}{0.8510}$ & $\underset{\text{\tiny(0.0118)}}{0.8444}$ & $\underset{\text{\tiny(0.0037)}}{0.8627}$ \\
            \textsc{CT-BERT} & $\underset{\text{\tiny(0.0167)}}{\textbf{0.8755}}$ & $\underset{\text{\tiny(0.0048)}}{0.8310}$ & $\underset{\text{\tiny(0.0131)}}{0.8258}$ & $\underset{\text{\tiny(0.0085)}}{0.8401}$ & $\underset{\text{\tiny(0.0194)}}{0.8232}$ & $\underset{\text{\tiny(0.0062)}}{0.8345}$ & $\underset{\text{\tiny(0.0109)}}{0.8280}$ & $\underset{\text{\tiny(0.0155)}}{0.8413}$ \\
            \textsc{RoBERTa-Base} & $\underset{\text{\tiny(0.0059)}}{\textbf{0.8986}}$ & $\underset{\text{\tiny(0.0173)}}{0.8572}$ & $\underset{\text{\tiny(0.0116)}}{0.8513}$ & $\underset{\text{\tiny(0.0092)}}{0.8669}$ & $\underset{\text{\tiny(0.0141)}}{0.8490}$ & $\underset{\text{\tiny(0.0035)}}{0.8608}$ & $\underset{\text{\tiny(0.0188)}}{0.8541}$ & $\underset{\text{\tiny(0.0079)}}{0.8876}$ \\
            \bottomrule
            \multicolumn{9}{l}{All model performance is evaluated using the F1-score.}\\
            \multicolumn{9}{l}{$^1$ The best results are highlighted in \textbf{bold}.}\\
            \multicolumn{9}{l}{$^2$ For deep learning models, results are reported as the average across 5-fold cross-validation, with standard deviation shown in parentheses.}
        \end{tabular}
    }
\end{table*}

\subsection{Sensitivity Analysis on Training Data Scale}

\begin{figure}[htbp]
    \centering

    \begin{minipage}{0.48\textwidth}
        \centering
        \includegraphics[width=\linewidth]{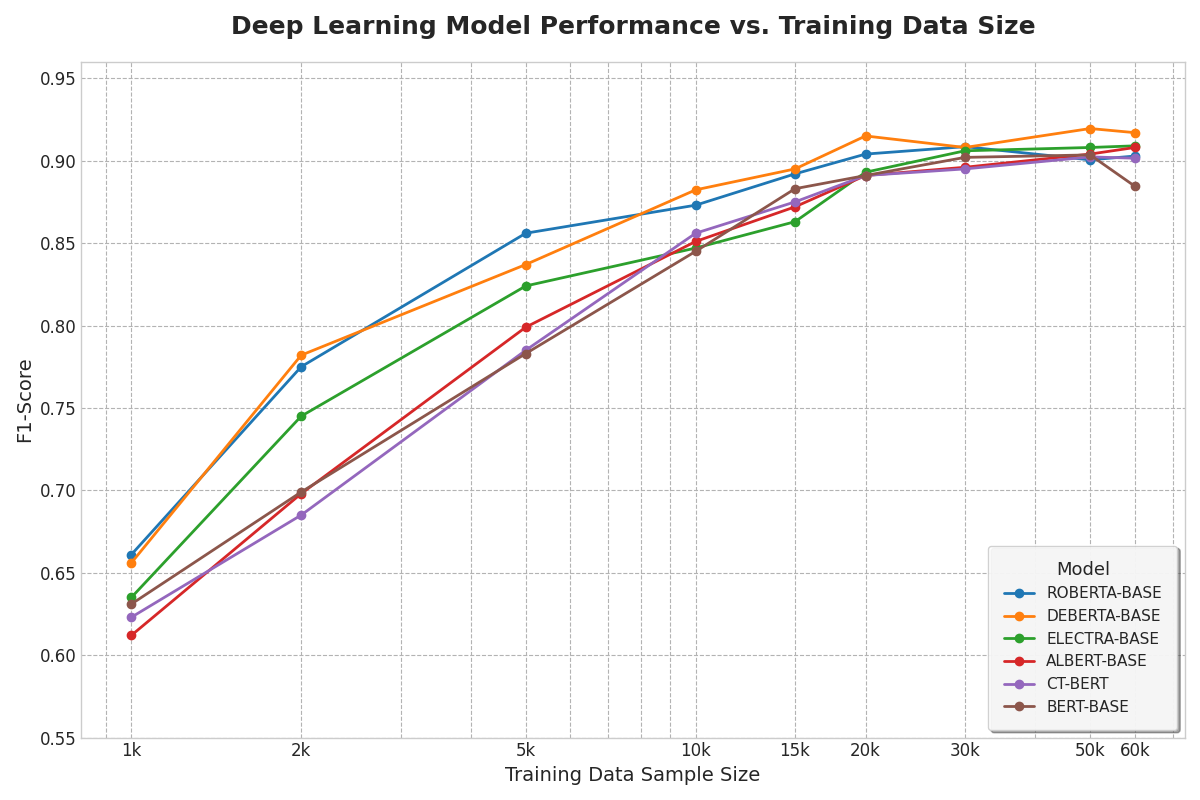}
        \label{fig:left_image}
    \end{minipage}
    \hfill 
    \begin{minipage}{0.48\textwidth}
        \centering
        \includegraphics[width=\linewidth]{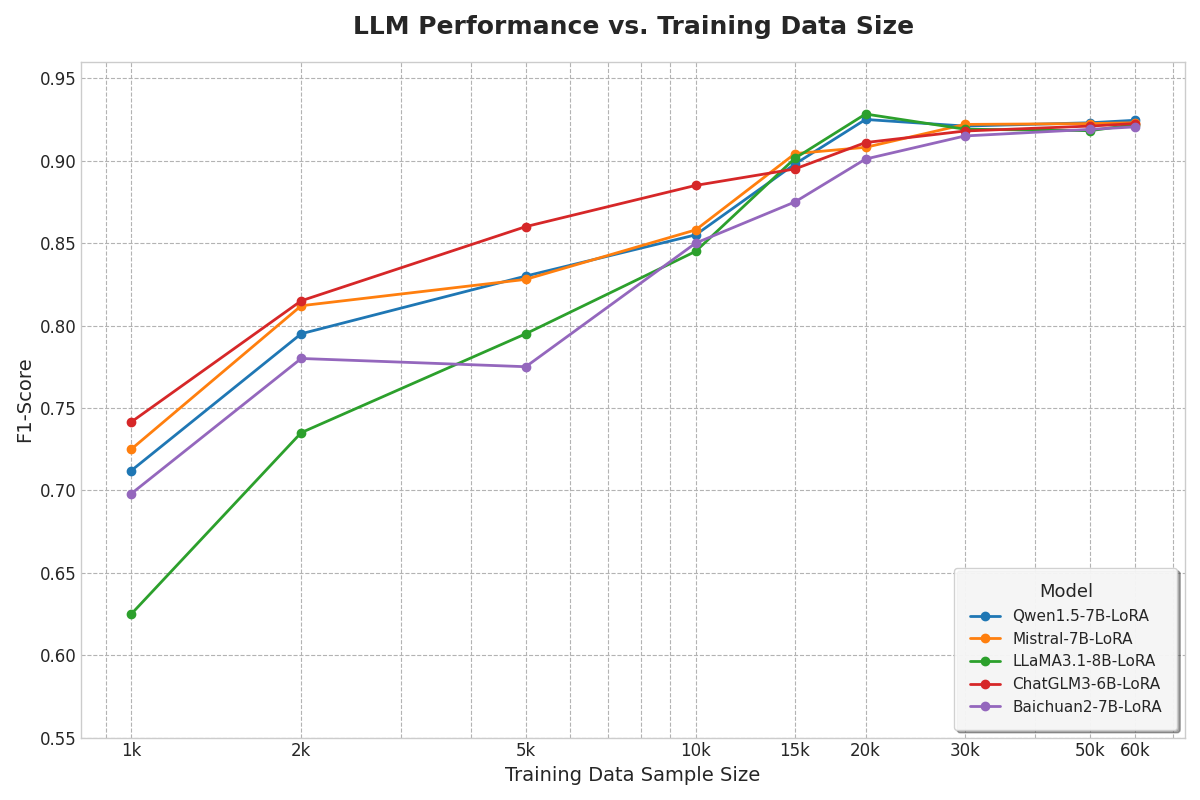}
        \label{fig:right_image}
    \end{minipage}
    
    \caption{Sensitivity Analysis on Training Data Scale}
    \label{fig:both_images}
\end{figure}

To empirically assess the impact of dataset size on detection performance and validate the necessity of MegaFake's scale, we conduct a sensitivity analysis using representative state-of-the-art models. We train both deep learning models and LLMs on MegaFake subsets ranging from 1k to 60k instances (specifically: 1k, 2k, 5k, 10k, 15k, 20k, 30k, 50k, and 60k) and test them on the remaining data. The results, illustrated in \textbf{Figure \ref{fig:both_images}}, reveal several critical patterns in the scaling law of deepfake text detection. We observe a steep performance increase between 1k and 20k training samples. For example, the F1-score of DeBERTa-Base rises from approximately 0.65 (at 1k) to over 0.90 (at 20k). Similarly, LLMs such as Qwen1.5-7B and LLaMA3.1-8B rapidly converge from 0.62–0.71 to around 0.92 over the same interval. This trend highlights the limitations of existing machine-generated fake news datasets, which typically include only a few thousand samples. Models trained on such small datasets are very likely under-optimized and fail to approach their performance ceiling. By contrast, MegaFake can provide the volume needed to traverse this steep part of the learning curve, enabling state-of-the-art models to realize their full detection potential.

Beyond the 20k sample threshold, the performance gains begins to plateau, exhibiting diminishing marginal returns. While F1-scores continue to edge upward or stabilize around 90\%–92\% as the training size increases to 50k and beyond, the rate of improvement is markedly slower than in the initial phase. This saturation point suggests that roughly 20k samples are sufficient to capture the core linguistic patterns of machine-generated text, whereas the additional data in MegaFake helps ensure robust generalization and stability across models and test conditions. Among the deep learning models, DeBERTa-Base and RoBERTa-Base consistently demonstrate strong data efficiency and peak performance, both reaching F1-scores above 0.91. Older architectures like BERT-Base initially lag but eventually converged to competitive levels when provided with sufficient data. For LLMs, LLaMA3.1-8B and Qwen1.5-7B emerge as the top performers, showing notable adaptability even at lower data scales, whereas Baichuan2-7B requires more data to narrow the performance gap.

Overall, this sensitivity analysis shows that MegaFake's scale is not redundant but essential for training high-performance detection models that move beyond the constraints imposed by smaller existing datasets.

\section{Discussion and Conclusion}
In this paper, we present MegaFake, a large-scale, theory-guided dataset designed to advance research in machine-generated fake news detection and governance. MegaFake is constructed from the GossipCop and PolitiFact datasets \citep{shu2020fakenewsnet} using both GLM \citep{glm2024chatglm} and Llama \citep{touvron2023llama}, comprising 129,085 fake news and 41,706 legitimate news instances for each model. Through a series of experiments, we demonstrate the utility of MegaFake in evaluating and benchmarking detection models across both human- and machine-generated news. By establishing rigorous benchmarks and enabling cross-domain analysis, MegaFake provides an essential foundation for developing more robust, adaptable, and transparent detection methods for supporting more effective online content and deepfake text governance.

\subsection{Theoretical and Practical Implications}
Our work provides several important theoretical and practical implications. Theoretically, we advance the understanding of digital deception by systematically analyzing both human- and machine-generated fake news. The introduction of the MegaFake dataset and the LLM-Fake Theory extends existing taxonomies of fake news, compelling researchers to reconsider traditional frameworks to accommodate the unique characteristics of machine-generated content. Our cross-domain experiments reveal important limitations in the generalizability of current detection models, suggesting the need for more adaptive and comprehensive theoretical models that can capture the evolving landscape of misinformation. In providing a systematic approach to generating and categorizing machine-generated fake news, our research lays a foundation for future theoretical studies on the mechanisms and impacts of AI-driven misinformation.

Practically, MegaFake serves as a robust, large-scale benchmark for evaluating and improving fake news detection systems, particularly those aimed at identifying deepfake text. The public availability of this dataset facilitates reproducibility and innovation, supporting both academic research and industry applications. Insights from our experiments can guide social media platforms, policymakers, and AI practitioners in developing more resilient moderation strategies and content governance policies to address the challenges posed by machine-generated misinformation. In particular, given the rapidly evolving nature of deception in the age of generative AI, our dataset can enhance existing fake news detection models by offering a diverse and up-to-date collection of sophisticated fake news examples that reflect current generative AI capabilities. This enables models to better detect emerging patterns of deception, improve their generalization to novel threats, and remain robust against increasingly advanced fake news tactics.

Furthermore, we address the evolving role of dataset scale in the era of LLMs. While recent advancements allow LLMs to be fine-tuned on relatively smaller datasets, the scale of MegaFake remains a critical asset for several reasons. First, a large and diverse dataset is essential for ensuring robustness and generalization, reducing the risks that models overfit to narrow patterns present in small samples. Second, the scale enables the construction of large, stable test sets, providing reliable benchmarks with low statistical variance. Finally, the volume of data supports fine-grained analysis of specific fake news categories and enables sensitivity studies on data scaling—analyses that are infeasible with limited datasets.

\subsection{Limitations and Future Directions}
Our study offers several important implications for future research. First, regarding domain coverage and generalizability, while MegaFake integrates political and entertainment news via PolitiFact and GossipCop, it does not yet encompass specialized domains such as health, finance, or science. Misinformation in these high-stakes fields often involves distinct linguistic patterns, technical jargon, and specific emotional triggers that may differ from general political or celebrity rumors. Consequently, detection models trained solely on MegaFake might experience performance degradation when applied to these specialized contexts. Future research should prioritize extending MegaFake to include domain-specific datasets and exploring cross-lingual misinformation to address the global nature of AI-generated content. Furthermore, MegaFake represents a crucial initial step toward creating datasets composed of machine-generated fake news. Future work may broaden the dataset’s scope by integrating a wider array of data sources. In addition, future research can diversify the prompts used to generate fake news. Exploring different prompt structures could yield variations in the qualities and characteristics of the generated fake news. Finally, future research can also seek to expand the LLM-Fake Theory to more comprehensively capture the multifaceted nature of fake news.

\begin{singlespace}
\small
\bibliographystyle{apalike}
\bibliography{custom}

@article{ng2021effect,
  title={The Effect of Platform Intervention Policies on Fake News Dissemination and Survival: An Empirical Examination},
  author={Ng, Ka Chung and Tang, Jie and Lee, Dongwon},
  journal={Journal of Management Information Systems},
  volume={38},
  number={4},
  pages={898--930},
  year={2021},
  publisher={Taylor \& Francis}
}

@misc{ma2026dearfinegrainedvlmadaptation,
      title={DeAR: Fine-Grained VLM Adaptation by Decomposing Attention Head Roles}, 
      author={Yiming Ma and Hongkun Yang and Lionel Z. Wang and Bin Chen and Weizhi Xian and Jianzhi Teng},
      year={2026},
      eprint={2603.01111},
      archivePrefix={arXiv},
      primaryClass={cs.CV},
      url={https://arxiv.org/abs/2603.01111}, 
}

@inproceedings{liu-etal-2025-sara,
    title = "{SARA}: Salience-Aware Reinforced Adaptive Decoding for Large Language Models in Abstractive Summarization",
    author = "Liu, Nayu  and
      Zhu, Junnan  and
      Ma, Yiming  and
      Lu, Zhicong  and
      Xu, Wenlei  and
      Yang, Yong  and
      Zhong, Jiang  and
      Wei, Kaiwen",
    editor = "Che, Wanxiang  and
      Nabende, Joyce  and
      Shutova, Ekaterina  and
      Pilehvar, Mohammad Taher",
    booktitle = "Proceedings of the 63rd Annual Meeting of the Association for Computational Linguistics (Volume 1: Long Papers)",
    month = jul,
    year = "2025",
    address = "Vienna, Austria",
    publisher = "Association for Computational Linguistics",
    url = "https://aclanthology.org/2025.acl-long.1236/",
    doi = "10.18653/v1/2025.acl-long.1236",
    pages = "25450--25463",
    ISBN = "979-8-89176-251-0"
}

@article{luo2023chatgpt,
  title={Chatgpt as a factual inconsistency evaluator for abstractive text summarization},
  author={Luo, Zheheng and Xie, Qianqian and Ananiadou, Sophia},
  journal={arXiv preprint arXiv:2303.15621},
  year={2023}
}

@article{ROSSI2024102749,
title = {Augmenting research methods with foundation models and generative AI},
journal = {International Journal of Information Management},
volume = {77},
pages = {102749},
year = {2024},
issn = {0268-4012},
doi = {https://doi.org/10.1016/j.ijinfomgt.2023.102749},
url = {https://www.sciencedirect.com/science/article/pii/S0268401223001305},
author = {Sippo Rossi and Matti Rossi and Raghava Rao Mukkamala and Jason Bennett Thatcher and Yogesh K. Dwivedi}
}

@inproceedings{wang-2017-liar,
    title = "``Liar, Liar Pants on Fire'': A New Benchmark Dataset for Fake News Detection",
    author = "Wang, William Yang",
    editor = "Barzilay, Regina  and
      Kan, Min-Yen",
    booktitle = "Proceedings of the 55th Annual Meeting of the Association for Computational Linguistics (Volume 2: Short Papers)",
    month = jul,
    year = "2017",
    address = "Vancouver, Canada",
    publisher = "Association for Computational Linguistics",
    url = "https://aclanthology.org/P17-2067/",
    doi = "10.18653/v1/P17-2067",
    pages = "422--426"
}

@article{susarla2023janus,
  title={The Janus effect of generative AI: Charting the path for responsible conduct of scholarly activities in information systems},
  author={Susarla, Anjana and Gopal, Ram and Thatcher, Jason Bennett and Sarker, Suprateek},
  journal={Information Systems Research},
  volume={34},
  number={2},
  pages={399--408},
  year={2023},
  publisher={INFORMS}
}

@article{10.1145/3137597.3137600,
author = {Shu, Kai and Sliva, Amy and Wang, Suhang and Tang, Jiliang and Liu, Huan},
title = {Fake News Detection on Social Media: A Data Mining Perspective},
year = {2017},
issue_date = {June 2017},
publisher = {Association for Computing Machinery},
address = {New York, NY, USA},
volume = {19},
number = {1},
issn = {1931-0145},
url = {https://doi.org/10.1145/3137597.3137600},
doi = {10.1145/3137597.3137600},
journal = {SIGKDD Explor. Newsl.},
month = sep,
pages = {22–36},
numpages = {15}
}

@article{doi:10.1126/science.aap9559,
author = {Soroush Vosoughi  and Deb Roy  and Sinan Aral },
title = {The spread of true and false news online},
journal = {Science},
volume = {359},
number = {6380},
pages = {1146-1151},
year = {2018},
doi = {10.1126/science.aap9559},
URL = {https://www.science.org/doi/abs/10.1126/science.aap9559},
eprint = {https://www.science.org/doi/pdf/10.1126/science.aap9559}}

@article{kaliyar2021fakebert,
  title={FakeBERT: Fake news detection in social media with a BERT-based deep learning approach},
  author={Kaliyar, Rohit Kumar and Goswami, Anurag and Narang, Pratik},
  journal={Multimedia tools and applications},
  volume={80},
  number={8},
  pages={11765--11788},
  year={2021},
  publisher={Springer}
}

@article{shu2020fakenewsnet,
  title={Fakenewsnet: A data repository with news content, social context, and spatiotemporal information for studying fake news on social media},
  author={Shu, Kai and Mahudeswaran, Deepak and Wang, Suhang and Lee, Dongwon and Liu, Huan},
  journal={Big data},
  volume={8},
  number={3},
  pages={171--188},
  year={2020},
  publisher={Mary Ann Liebert, Inc., publishers 140 Huguenot Street, 3rd Floor New~…}
}

@inproceedings{chen2024llmgenerated,
      title={Can {LLM}-Generated Misinformation Be Detected?},
      author={Canyu Chen and Kai Shu},
      booktitle={The Twelfth International Conference on Learning Representations},
      year={2024},
      url={https://openreview.net/forum?id=ccxD4mtkTU}
      }

@article{schuster2020limitations,
  title={The limitations of stylometry for detecting machine-generated fake news},
  author={Schuster, Tal and Schuster, Roei and Shah, Darsh J and Barzilay, Regina},
  journal={Computational Linguistics},
  volume={46},
  number={2},
  pages={499--510},
  year={2020},
  publisher={MIT Press One Rogers Street, Cambridge, MA 02142-1209, USA journals-info~…}
}

@article{zhang2024benchmarking,
  title={Benchmarking large language models for news summarization},
  author={Zhang, Tianyi and Ladhak, Faisal and Durmus, Esin and Liang, Percy and McKeown, Kathleen and Hashimoto, Tatsunori B},
  journal={Transactions of the Association for Computational Linguistics},
  volume={12},
  pages={39--57},
  year={2024},
  publisher={MIT Press One Broadway, 12th Floor, Cambridge, Massachusetts 02142, USA~…}
}

@article{murayama2021dataset,
  title={Dataset of fake news detection and fact verification: a survey},
  author={Murayama, Taichi},
  journal={arXiv preprint arXiv:2111.03299},
  year={2021}
}

@article{bai2023qwen,
  title={Qwen technical report},
  author={Bai, Jinze and Bai, Shuai and Chu, Yunfei and Cui, Zeyu and Dang, Kai and Deng, Xiaodong and Fan, Yang and Ge, Wenbin and Han, Yu and Huang, Fei and others},
  journal={arXiv preprint arXiv:2309.16609},
  year={2023}
}

@inproceedings{jiao-etal-2020-tinybert,
    title = "{T}iny{BERT}: Distilling {BERT} for Natural Language Understanding",
    author = "Jiao, Xiaoqi  and
      Yin, Yichun  and
      Shang, Lifeng  and
      Jiang, Xin  and
      Chen, Xiao  and
      Li, Linlin  and
      Wang, Fang  and
      Liu, Qun",
    editor = "Cohn, Trevor  and
      He, Yulan  and
      Liu, Yang",
    booktitle = "Findings of the Association for Computational Linguistics: EMNLP 2020",
    month = nov,
    year = "2020",
    address = "Online",
    publisher = "Association for Computational Linguistics",
    url = "https://aclanthology.org/2020.findings-emnlp.372/",
    doi = "10.18653/v1/2020.findings-emnlp.372",
    pages = "4163--4174"
}

@misc{huang2024fakegpt,
      title={FakeGPT: Fake News Generation, Explanation and Detection of Large Language Models}, 
      author={Yue Huang and Lichao Sun},
      year={2024},
      eprint={2310.05046},
      archivePrefix={arXiv},
      primaryClass={cs.CL}
}

@article{muller2023covid,
  title={Covid-twitter-bert: A natural language processing model to analyse covid-19 content on twitter},
  author={M{\"u}ller, Martin and Salath{\'e}, Marcel and Kummervold, Per E},
  journal={Frontiers in artificial intelligence},
  volume={6},
  pages={1023281},
  year={2023},
  publisher={Frontiers Media SA}
}

@inproceedings{lan2019albert,
  title={Albert: A lite bert for self-supervised learning of language representations},
  author={Lan, Zhenzhong and Chen, Mingda and Goodman, Sebastian and Gimpel, Kevin and Sharma, Piyush and Soricut, Radu},
  booktitle={The Eighth International Conference on Learning Representations},
  year={2019}
}

@article{liu2019roberta,
  title={Roberta: A robustly optimized bert pretraining approach},
  author={Liu, Yinhan and Ott, Myle and Goyal, Naman and Du, Jingfei and Joshi, Mandar and Chen, Danqi and Levy, Omer and Lewis, Mike and Zettlemoyer, Luke and Stoyanov, Veselin},
  journal={arXiv preprint arXiv:1907.11692},
  year={2019}
}

@inproceedings{alambo2020topic,
  title={Topic-centric unsupervised multi-document summarization of scientific and news articles},
  author={Alambo, Amanuel and Lohstroh, Cori and Madaus, Erik and Padhee, Swati and Foster, Brandy and Banerjee, Tanvi and Thirunarayan, Krishnaprasad and Raymer, Michael},
  booktitle={2020 IEEE International Conference on Big Data (Big Data)},
  pages={591--596},
  year={2020},
  organization={IEEE}
}

@article{vayansky2020review,
  title={A review of topic modeling methods},
  author={Vayansky, Ike and Kumar, Sathish AP},
  journal={Information Systems},
  volume={94},
  pages={101582},
  year={2020},
  publisher={Elsevier}
}

@article{xu2023chatgpt,
  title={ChatGPT vs. Google: a comparative study of search performance and user experience},
  author={Xu, Ruiyun and Feng, Yue and Chen, Hailiang},
  journal={arXiv preprint arXiv:2307.01135},
  year={2023}
}

@inproceedings{du2023glm,
  author    = {Zhengxiao Du and
               Yujie Qian and
               Xiao Liu and
               Ming Ding and
               Jiezhong Qiu and
               Zhilin Yang and
               Jie Tang},
  title     = {{GLM:} General Language Model Pretraining with Autoregressive Blank Infilling},
  booktitle = {Proceedings of the 60th Annual Meeting of the Association for Computational
               Linguistics (Volume 1: Long Papers), {ACL} 2022, Dublin, Ireland,
               May 22-27, 2022},
  pages     = {320--335},
  publisher = {Association for Computational Linguistics},
  year      = {2022},
}

@article{yang2023chatgpt,
  title={Is chatgpt involved in texts? measure the polish ratio to detect chatgpt-generated text},
  author={Yang, Lingyi and Jiang, Feng and Li, Haizhou and others},
  journal={APSIPA Transactions on Signal and Information Processing},
  volume={13},
  number={2},
  year={2023},
  publisher={Now Publishers, Inc.}
}

@article{chatman1975towards,
  title={Towards a theory of narrative},
  author={Chatman, Seymour},
  journal={New literary history},
  volume={6},
  number={2},
  pages={295--318},
  year={1975},
  publisher={JSTOR}
}

@article{bhattacherjee2006influence,
  title={Influence processes for information technology acceptance: An elaboration likelihood model},
  author={Bhattacherjee, Anol and Sanford, Clive},
  journal={MIS quarterly},
  pages={805--825},
  year={2006},
  publisher={JSTOR}
}

@article{lazer2018science,
  title={The science of fake news},
  author={Lazer, David MJ and Baum, Matthew A and Benkler, Yochai and Berinsky, Adam J and Greenhill, Kelly M and Menczer, Filippo and Metzger, Miriam J and Nyhan, Brendan and Pennycook, Gordon and Rothschild, David and others},
  journal={Science},
  volume={359},
  number={6380},
  pages={1094--1096},
  year={2018},
  publisher={American Association for the Advancement of Science}
}

@article{moravec2018fake,
  title={Fake news on social media: People believe what they want to believe when it makes no sense at all},
  author={Moravec, Patricia and Minas, Randall and Dennis, Alan R},
  journal={Kelley School of Business research paper},
  year={2018}
}

@article{hinojosa2017review,
  title={A review of cognitive dissonance theory in management research: Opportunities for further development},
  author={Hinojosa, Amanda S and Gardner, William L and Walker, H Jack and Cogliser, Claudia and Gullifor, Daniel},
  journal={Journal of Management},
  volume={43},
  number={1},
  pages={170--199},
  year={2017},
  publisher={Sage Publications Sage CA: Los Angeles, CA}
}

@article{petty2011elaboration,
  title={The elaboration likelihood model},
  author={Petty, Richard E and Bri{\~n}ol, Pablo},
  journal={Handbook of theories of social psychology},
  volume={1},
  pages={224--245},
  year={2011},
  publisher={Sage Publications Ltd Los Angeles}
}

@article{chen2020linguistic,
  title={A linguistic signaling model of social support exchange in online health communities},
  author={Chen, Langtao and Baird, Aaron and Straub, Detmar},
  journal={Decision Support Systems},
  volume={130},
  pages={113233},
  year={2020},
  publisher={Elsevier}
}

@inproceedings{zhou2023synthetic,
  title={Synthetic lies: Understanding ai-generated misinformation and evaluating algorithmic and human solutions},
  author={Zhou, Jiawei and Zhang, Yixuan and Luo, Qianni and Parker, Andrea G and De Choudhury, Munmun},
  booktitle={Proceedings of the 2023 CHI Conference on Human Factors in Computing Systems},
  pages={1--20},
  year={2023}
}

@inproceedings{jiang2024disinformation,
  title={Disinformation detection: An evolving challenge in the age of llms},
  author={Jiang, Bohan and Tan, Zhen and Nirmal, Ayushi and Liu, Huan},
  booktitle={Proceedings of the 2024 SIAM International Conference on Data Mining (SDM)},
  pages={427--435},
  year={2024},
  organization={SIAM}
}

@inproceedings{lucas-etal-2023-fighting,
    title = "Fighting Fire with Fire: The Dual Role of {LLM}s in Crafting and Detecting Elusive Disinformation",
    author = "Lucas, Jason  and
      Uchendu, Adaku  and
      Yamashita, Michiharu  and
      Lee, Jooyoung  and
      Rohatgi, Shaurya  and
      Lee, Dongwon",
    editor = "Bouamor, Houda  and
      Pino, Juan  and
      Bali, Kalika",
    booktitle = "Proceedings of the 2023 Conference on Empirical Methods in Natural Language Processing",
    month = dec,
    year = "2023",
    address = "Singapore",
    publisher = "Association for Computational Linguistics",
    url = "https://aclanthology.org/2023.emnlp-main.883/",
    doi = "10.18653/v1/2023.emnlp-main.883",
    pages = "14279--14305"
}

@inproceedings{vykopal-etal-2024-disinformation,
    title = "Disinformation Capabilities of Large Language Models",
    author = "Vykopal, Ivan  and
      Pikuliak, Mat{\'u}{\v{s}}  and
      Srba, Ivan  and
      Moro, Robert  and
      Macko, Dominik  and
      Bielikova, Maria",
    editor = "Ku, Lun-Wei  and
      Martins, Andre  and
      Srikumar, Vivek",
    booktitle = "Proceedings of the 62nd Annual Meeting of the Association for Computational Linguistics (Volume 1: Long Papers)",
    month = aug,
    year = "2024",
    address = "Bangkok, Thailand",
    publisher = "Association for Computational Linguistics",
    url = "https://aclanthology.org/2024.acl-long.793/",
    doi = "10.18653/v1/2024.acl-long.793",
    pages = "14830--14847"
}

@article{10.1145/3703155,
author = {Huang, Lei and Yu, Weijiang and Ma, Weitao and Zhong, Weihong and Feng, Zhangyin and Wang, Haotian and Chen, Qianglong and Peng, Weihua and Feng, Xiaocheng and Qin, Bing and Liu, Ting},
title = {A Survey on Hallucination in Large Language Models: Principles, Taxonomy, Challenges, and Open Questions},
year = {2025},
issue_date = {March 2025},
publisher = {Association for Computing Machinery},
address = {New York, NY, USA},
volume = {43},
number = {2},
issn = {1046-8188},
url = {https://doi.org/10.1145/3703155},
doi = {10.1145/3703155},
journal = {ACM Trans. Inf. Syst.},
month = jan,
articleno = {42},
numpages = {55}
}

@article{zhang2023siren,
  title={Siren's song in the AI ocean: a survey on hallucination in large language models},
  author={Zhang, Yue and Li, Yafu and Cui, Leyang and Cai, Deng and Liu, Lemao and Fu, Tingchen and Huang, Xinting and Zhao, Enbo and Zhang, Yu and Chen, Yulong and others},
  journal={arXiv preprint arXiv:2309.01219},
  year={2023}
}

@inproceedings{hu2024bad,
  title={Bad actor, good advisor: Exploring the role of large language models in fake news detection},
  author={Hu, Beizhe and Sheng, Qiang and Cao, Juan and Shi, Yuhui and Li, Yang and Wang, Danding and Qi, Peng},
  booktitle={Proceedings of the AAAI Conference on Artificial Intelligence},
  volume={38},
  pages={22105--22113},
  year={2024}
}

@inproceedings{wang2024explainable,
  title={Explainable Fake News Detection With Large Language Model via Defense Among Competing Wisdom},
  author={Wang, Bo and Ma, Jing and Lin, Hongzhan and Yang, Zhiwei and Yang, Ruichao and Tian, Yuan and Chang, Yi},
  booktitle={Proceedings of the ACM on Web Conference 2024},
  pages={2452--2463},
  year={2024}
}

@article{glm2024chatglm,
  title={ChatGLM: A Family of Large Language Models from GLM-130B to GLM-4 All Tools},
  author={GLM, Team and Zeng, Aohan and Xu, Bin and Wang, Bowen and Zhang, Chenhui and Yin, Da and Rojas, Diego and Feng, Guanyu and Zhao, Hanlin and Lai, Hanyu and others},
  journal={arXiv preprint arXiv:2406.12793},
  year={2024}
}

@article{achiam2023gpt,
  title={Gpt-4 technical report},
  author={Achiam, Josh and Adler, Steven and Agarwal, Sandhini and Ahmad, Lama and Akkaya, Ilge and Aleman, Florencia Leoni and Almeida, Diogo and Altenschmidt, Janko and Altman, Sam and Anadkat, Shyamal and others},
  journal={arXiv preprint arXiv:2303.08774},
  year={2023}
}

@article{touvron2023llama,
  title={Llama: Open and efficient foundation language models},
  author={Touvron, Hugo and Lavril, Thibaut and Izacard, Gautier and Martinet, Xavier and Lachaux, Marie-Anne and Lacroix, Timoth{\'e}e and Rozi{\`e}re, Baptiste and Goyal, Naman and Hambro, Eric and Azhar, Faisal and others},
  journal={arXiv preprint arXiv:2302.13971},
  year={2023}
}

@article{jiang2023mistral,
  title={Mistral 7B},
  author={Jiang, Albert Q and Sablayrolles, Alexandre and Mensch, Arthur and Bamford, Chris and Chaplot, Devendra Singh and Casas, Diego de las and Bressand, Florian and Lengyel, Gianna and Lample, Guillaume and Saulnier, Lucile and others},
  journal={arXiv preprint arXiv:2310.06825},
  year={2023}
}

@article{hu2021lora,
  title={Lora: Low-rank adaptation of large language models},
  author={Hu, Edward J and Shen, Yelong and Wallis, Phillip and Allen-Zhu, Zeyuan and Li, Yuanzhi and Wang, Shean and Wang, Lu and Chen, Weizhu},
  journal={The Tenth International Conference on Learning Representations (Virtual)},
  year={2021}
}

@article{yang2023baichuan,
  title={Baichuan 2: Open large-scale language models},
  author={Yang, Aiyuan and Xiao, Bin and Wang, Bingning and Zhang, Borong and Bian, Ce and Yin, Chao and Lv, Chenxu and Pan, Da and Wang, Dian and Yan, Dong and others},
  journal={arXiv preprint arXiv:2309.10305},
  year={2023}
}

@article{pennycook2021psychology,
  title={The psychology of fake news},
  author={Pennycook, Gordon and Rand, David G},
  journal={Trends in cognitive sciences},
  volume={25},
  number={5},
  pages={388--402},
  year={2021},
  publisher={Elsevier}
}

@article{zhang2020overview,
  title={An overview of online fake news: Characterization, detection, and discussion},
  author={Zhang, Xichen and Ghorbani, Ali A},
  journal={Information Processing \& Management},
  volume={57},
  number={2},
  pages={102025},
  year={2020},
  publisher={Elsevier}
}

@article{gelfert2018fake,
  title={Fake news: A definition},
  author={Gelfert, Axel},
  journal={Informal logic},
  volume={38},
  number={1},
  pages={84--117},
  year={2018},
  publisher={{\'E}rudit}
}

@article{lee2024explainable,
  title={Explainable deep learning for false information identification: An argumentation theory approach},
  author={Lee, Kyuhan and Ram, Sudha},
  journal={Information Systems Research},
  volume={35},
  number={2},
  pages={890--907},
  year={2024},
  publisher={INFORMS}
}

@article{ng2023augmenting,
  title={Augmenting fake content detection in online platforms: A domain adaptive transfer learning via adversarial training approach},
  author={Ng, Ka Chung and Ke, Ping Fan and So, Mike KP and Tam, Kar Yan},
  journal={Production and Operations Management},
  volume={32},
  number={7},
  pages={2101--2122},
  year={2023},
  publisher={SAGE Publications Sage CA: Los Angeles, CA}
}

@article{zhang2022theory,
  title={A theory-driven machine learning system for financial disinformation detection},
  author={Zhang, Xiaohui and Du, Qianzhou and Zhang, Zhongju},
  journal={Production and Operations Management},
  volume={31},
  number={8},
  pages={3160--3179},
  year={2022},
  publisher={SAGE Publications Sage CA: Los Angeles, CA}
}

@article{zhou2020fake,
  title={Fake news early detection: A theory-driven model},
  author={Zhou, Xinyi and Jain, Atishay and Phoha, Vir V and Zafarani, Reza},
  journal={Digital Threats: Research and Practice},
  volume={1},
  number={2},
  pages={1--25},
  year={2020},
  publisher={ACM New York, NY, USA}
}

@inproceedings{shu2019beyond,
  title={Beyond news contents: The role of social context for fake news detection},
  author={Shu, Kai and Wang, Suhang and Liu, Huan},
  booktitle={Proceedings of the twelfth ACM international conference on web search and data mining},
  pages={312--320},
  year={2019}
}

@inproceedings{pan2018content,
  title={Content based fake news detection using knowledge graphs},
  author={Pan, Jeff Z and Pavlova, Siyana and Li, Chenxi and Li, Ningxi and Li, Yangmei and Liu, Jinshuo},
  booktitle={The Semantic Web--ISWC 2018: 17th International Semantic Web Conference, Monterey, CA, USA, October 8--12, 2018, Proceedings, Part I 17},
  pages={669--683},
  year={2018},
  organization={Springer}
}

@inproceedings{della2018automatic,
  title={Automatic online fake news detection combining content and social signals},
  author={Della Vedova, Marco L and Tacchini, Eugenio and Moret, Stefano and Ballarin, Gabriele and DiPierro, Massimo and De Alfaro, Luca},
  booktitle={2018 22nd conference of open innovations association (FRUCT)},
  pages={272--279},
  year={2018},
  organization={IEEE}
}

@article{grace2018will,
  title={When will AI exceed human performance? Evidence from AI experts},
  author={Grace, Katja and Salvatier, John and Dafoe, Allan and Zhang, Baobao and Evans, Owain},
  journal={Journal of Artificial Intelligence Research},
  volume={62},
  pages={729--754},
  year={2018}
}

@article{korteling2021human,
  title={Human-versus artificial intelligence},
  author={Korteling, JE (Hans) and van de Boer-Visschedijk, Gillian C and Blankendaal, Romy AM and Boonekamp, Rudy C and Eikelboom, Aletta R},
  journal={Frontiers in artificial intelligence},
  volume={4},
  pages={622364},
  year={2021},
  publisher={Frontiers Media SA}
}

@article{clark2020electra,
  title={Electra: Pre-training text encoders as discriminators rather than generators},
  author={Clark, Kevin and Luong, Minh-Thang and Le, Quoc V and Manning, Christopher D},
  journal={The Eighth International Conference on Learning Representations},
  year={2020}
}

@article{he2020deberta,
  title={Deberta: Decoding-enhanced bert with disentangled attention},
  author={He, Pengcheng and Liu, Xiaodong and Gao, Jianfeng and Chen, Weizhu},
  journal={The Ninth International Conference on Learning Representations},
  year={2020}
}

@inproceedings{shi-etal-2023-hallucination,
    title = "Hallucination Mitigation in Natural Language Generation from Large-Scale Open-Domain Knowledge Graphs",
    author = "Shi, Xiao  and
      Zhu, Zhengyuan  and
      Zhang, Zeyu  and
      Li, Chengkai",
    editor = "Bouamor, Houda  and
      Pino, Juan  and
      Bali, Kalika",
    booktitle = "Proceedings of the 2023 Conference on Empirical Methods in Natural Language Processing",
    month = dec,
    year = "2023",
    address = "Singapore",
    publisher = "Association for Computational Linguistics",
    url = "https://aclanthology.org/2023.emnlp-main.770/",
    doi = "10.18653/v1/2023.emnlp-main.770",
    pages = "12506--12521"
}

@inproceedings{su2024adapting,
  title={Adapting fake news detection to the era of large language models},
  author={Su, Jinyan and Cardie, Claire and Nakov, Preslav},
  booktitle={Findings of the association for computational linguistics: NAACL 2024},
  pages={1473--1490},
  year={2024}
}

@article{sun2024exploring,
  title={Exploring the deceptive power of llm-generated fake news: A study of real-world detection challenges},
  author={Sun, Yanshen and He, Jianfeng and Cui, Limeng and Lei, Shuo and Lu, Chang-Tien},
  journal={arXiv preprint arXiv:2403.18249},
  year={2024}
}

@inproceedings{wu2024sheepdog,
author = {Wu, Jiaying and Guo, Jiafeng and Hooi, Bryan},
title = {Fake News in Sheep's Clothing: Robust Fake News Detection Against LLM-Empowered Style Attacks},
year = {2024},
booktitle = {Proceedings of the 30th ACM SIGKDD Conference on Knowledge Discovery and Data Mining},
pages = {3367–3378}
}

@article{das2025fake,
  title={Fake news detection after llm laundering: Measurement and explanation},
  author={Das, Rupak Kumar and Dodge, Jonathan},
  journal={arXiv preprint arXiv:2501.18649},
  year={2025}
}

@article{wang2025prompt,
  title={Prompt-Induced Linguistic Fingerprints for LLM-Generated Fake News Detection},
  author={Wang, Chi and Gao, Min and Wang, Zongwei and Yin, Junwei and Shu, Kai and Lin, Chenghua},
  journal={arXiv preprint arXiv:2508.12632},
  year={2025}
}

@inproceedings{ayoobi2024seeing,
  title={Seeing Through AI's Lens: Enhancing Human Skepticism Towards LLM-Generated Fake News},
  author={Ayoobi, Navid and Shahriar, Sadat and Mukherjee, Arjun},
  booktitle={Proceedings of the 35th ACM Conference on Hypertext and Social Media},
  pages={1--11},
  year={2024}
}

@inproceedings{hong2024leveraging,
  title={Leveraging LLMs for LLM-Generated Fake News Detection: Insights from COVID-19 Misinformation},
  author={Hong, Dao Ngoc and Hashimoto, Yasuhiro and Paik, Incheon and Thang, Truong Cong},
  booktitle={2024 IEEE 16th International Conference on Computational Intelligence and Communication Networks (CICN)},
  pages={1460--1466},
  year={2024},
  organization={IEEE}
}

@inproceedings{morita2025genaireading,
  title={GenAIReading: augmenting human cognition with interactive digital textbooks using large Language models and image generation models},
  author={Morita, Ryugo and Watanabe, Ko and Zhou, Jinjia and Dengel, Andreas and Ishimaru, Shoya},
  booktitle={Proceedings of the Augmented Humans International Conference 2025},
  pages={289--301},
  year={2025}
}

@inproceedings{watanabe2024comparing,
  title={Comparing Web Browsing Behaviors with High and Low Information Literacy: A Case Study for Fact Check Against GPT Generated Fake News},
  author={Watanabe, Ko and Tanaka, Seiya and Vargo, Andrew and Kise, Koichi and Dengel, Andreas},
  booktitle={Companion of the 2024 on ACM International Joint Conference on Pervasive and Ubiquitous Computing},
  pages={9--13},
  year={2024}
}

@article{solnyshkina2017evaluating,
  title={Evaluating text complexity and Flesch-Kincaid grade level},
  author={Solnyshkina, Marina and Zamaletdinov, Radif and Gorodetskaya, Ludmila and Gabitov, Azat},
  journal={Journal of social studies education research},
  volume={8},
  number={3},
  pages={238--248},
  year={2017},
  publisher={B{\"u}lent TARMAN}
}

@inproceedings{beasley2015emotional,
  title={Emotional states vs. emotional words in social media},
  author={Beasley, Asaf and Mason, Winter},
  booktitle={Proceedings of the ACM web science conference},
  pages={1--10},
  year={2015}
}
\end{singlespace}

\end{document}